\documentclass{article}
\pdfoutput=1
\usepackage{microtype}
\usepackage{graphicx}
\usepackage{subfigure}
\usepackage{booktabs} %

\usepackage{hyperref}

\usepackage{amsmath,amssymb,amsthm,amsfonts}
\usepackage{complexity}

\usepackage{xspace}
 
\newcommand{\CSS}{\texttt{BOCS}\xspace} 
\newcommand{\CSSSDP}{\texttt{BOCS-SDP}\xspace}
\newcommand{\CSSSA}{\texttt{BOCS-SA}\xspace}
\newcommand{\MLE}{\texttt{MLE}\xspace}
\newcommand{\EI}{\texttt{EI}\xspace}
\newcommand{\RS}{\texttt{RS}\xspace}
\newcommand{\SMAC}{\texttt{SMAC}\xspace}
\newcommand{\SA}{\texttt{SA}\xspace}
\newcommand{\OLS}{\texttt{OLS}\xspace}
\newcommand{\PS}{\texttt{PS}\xspace} %
\newcommand{\domain}{\ensuremath{{\cal D}}\xspace}

\usepackage[accepted]{icml2018}

\icmltitlerunning{Bayesian Optimization of Combinatorial Structures}

\newcommand{\balpha}{\boldsymbol{\alpha}}
\DeclareMathOperator*{\argmin}{argmin}
\DeclareMathOperator*{\argmax}{argmax}

\usepackage{tikz}
\usetikzlibrary{shapes,arrows,positioning}

\begin{document}

\twocolumn[
\icmltitle{Bayesian Optimization of Combinatorial Structures}

\icmlsetsymbol{equal}{*}

\begin{icmlauthorlist}
\icmlauthor{Ricardo Baptista}{mit}
\icmlauthor{Matthias Poloczek}{arizona}
\end{icmlauthorlist}

\icmlaffiliation{mit}{Center for Computational Engineering, Massachusetts Institute of Technology, Cambridge, MA}
\icmlaffiliation{arizona}{Dept.\ of Systems and Industrial Engineering, The University of Arizona, Tucson, AZ}

\icmlcorrespondingauthor{Ricardo Baptista}{rsb@mit.edu}
\icmlcorrespondingauthor{Matthias Poloczek}{poloczek@email.arizona.edu}

\icmlkeywords{Bayesian optimization, Combinatorial algorithms, Semidefinite optimization, Simulated annealing, Statistical model, Uncertainty quantification}

\vskip 0.3in
]

\printAffiliationsAndNotice{\icmlEqualContribution} %

\begin{abstract}
The optimization of expensive-to-evaluate black-box functions over combinatorial structures is an ubiquitous task in machine learning, engineering and the natural sciences. 
The combinatorial explosion of the search space and costly evaluations pose challenges for current techniques in discrete optimization and machine learning, and critically require new algorithmic ideas.
This article proposes, to the best of our knowledge, the first algorithm to overcome these challenges, based on an adaptive, scalable model that identifies useful combinatorial structure even when data is scarce.
Our acquisition function pioneers the use of semidefinite programming to achieve efficiency and scalability.
Experimental evaluations demonstrate that this algorithm consistently outperforms other methods from combinatorial and Bayesian optimization.
\end{abstract}

\section{Introduction} \label{sec:introduction}
We consider the problem of optimizing an expensive-to-evaluate black-box function over a set of combinatorial structures. 
This problem is pervasive in machine learning, engineering, and the natural sciences. 
Applications include object location in images~\cite{zhang_2015}, drug discovery~\cite{negoescu_2011}, cross-validation of hyper-parameters in mixed-integer solvers~\cite{hutter_2010}, food safety control~\cite{hu_2010}, and model sparsification in multi-component systems~\cite{baptista_2017}.
We also face this problem in the shared economy: a bike sharing company has to decide where to place bike stations among locations offered by the communal administration in order to optimize its utility, as measured in a field test.

We present a novel algorithm for this problem, \emph{Bayesian Optimization of Combinatorial Structures} (\CSS), that is capable of taming the combinatorial explosion of the search domain while achieving sample-efficiency, thereby improving substantially over the state of the art.  
Specifically, our contributions are:
\begin{enumerate}
\item A novel method to obtain an approximate optimizer of the acquisition function that employs algorithmic ideas from convex optimization to achieve scalability and efficiency. This approach overcomes the inherent limited scalability of many acquisition functions to large combinatorial domains.

\item We propose a model that captures the interaction of structural elements, and show how to infer these interactions in practice when data is expensive and thus scarce. 
We also demonstrate the usefulness of this interpretable model on experimental data.

\item We evaluate the performance of the \CSS algorithm together with methods from machine learning and discrete optimization on a variety of benchmark problems, including tasks from machine learning, aerospace engineering, and food safety control.

\end{enumerate}

\paragraph*{Related Work:}
Bayesian optimization has emerged as a powerful technique for the optimization of expensive functions if the domain is a box-constrained subset of the real coordinate space, i.e., a tensor-product of bounded connected univariate domains (e.g., see~\citet{brochu2010tutorial,shahriari_2016}). 
Recently, these techniques have been extended to certain high-dimensional problems whose box-constrained sets have a low `effective dimensionality'~\cite{wang2016bayesian,binois2017choice} or an additive decomposition~\cite{kandasamy2015high,li2016high,wang2017batched}. 
\citet{hutter2013kernel} and \citet{swersky2014raiders} considered applications where parameters have conditional dependencies, becoming irrelevant if other parameters take certain values.
\citet{jenatton2017bayesian} presented a scalable algorithm when these dependencies form a tree.
\citet{shahriari2016unbounded} proposed techniques for growing the box adaptively to optimize over unbounded continuous domains.

Structured domains have received little attention.
\citet{negoescu_2011} proposed a linear parametric belief model that can handle categorical variables. Their streamlined knowledge gradient acquisition function has cost~$\Omega(d\cdot 2^d)$ for each iteration and thus is designed for applications with small dimensionality~$d$.
\citet{hutter2011sequential} suggested a novel surrogate model based on random forests to handle categorical data. Their \SMAC algorithm uses random walks to obtain a local optimum under the expected improvement acquisition criterion \cite{MoTiZi78,jsw98}, and therefore can handle even high dimensional problems.
In practice, structured domains are often embedded into a box in~$\mathbb{R}^d$ to run an off-the-shelf Bayesian optimization software, e.g., see~\cite{dewancker_2016,golovin2017google}.
However, this is typically infeasible in practice due to the curse of dimensionality, also referred to as \emph{combinatorial explosion}, as the number of alternatives grows exponentially with the parameters. 
Thus, it is not surprising that optimization over structured domains was raised as an important open problem at the NIPS~2017 Workshop on Bayesian optimization~\cite{bayesopt_question}.

Methods in discrete optimization that are able to handle black-box functions include local search~\cite{kmsv98,skc94,spears_1993} and evolutionary algorithms, e.g., particle search~\cite{schafer_2013}. 
However, these procedures are not designed to be sample efficient and hence often prohibitively expensive for the problems we consider. 
Moreover, the popular local search algorithms have the conceptual disadvantage that they do not necessarily converge to a global optimum.
Popular techniques such as branch and bound and mathematical programming, e.g., linear, convex, and mixed-integer programming, typically cannot be applied to black-box functions.
We will compare the \CSS algorithm to the methods of~\cite{snoek_2012,hutter2011sequential,kmsv98,spears_1993,bb12,schafer_2013} in Sect.~\ref{sec:num_results}.

We formalize the problem under consideration in Sect.~\ref{section_problem},
describe the statistical model in Sect.~\ref{sec:statistical_model}, 
specify our acquisition function and the relaxation to semidefinite programming in Sects.~\ref{sec:acquisition_fn} and \ref{sec:css_alg}, 
present numerical experiments in Sect.~\ref{section_experiments},
and conclude in Sect.~\ref{section_conclusion}.
Sections labeled by letters are in the supplement.

The code for this paper is available at \url{https://github.com/baptistar/BOCS}.

\section{Problem Formulation}
\label{section_problem}
Given an expensive-to-evaluate black-box function~$f$ over a \emph{discrete structured domain}~$\domain$ of feasible points, our goal is to find a \emph{global optimizer}~$\argmax_{x \in \domain} f(x)$. 
We suppose that observing~$x$ provides independent, conditional on~$f(x)$, and normally distributed observations with mean~$f(x)$ and finite variance~$\sigma^2$. 
For the sake of simplicity, we focus on~$\domain = \{0,1\}^d$, where~$x_i$ equals one if a certain element~$i$ is present in the design and zero otherwise.  For example, we can associate a binary variable with each possible location for a bike station, with a side-chain in a chemical compound, with a possible coupling between two components in a multi-component system, or more generally with an edge in a graph-like structure.
We note that \CSS generalizes to integer-valued and categorical variables and to models of higher order (see Sect.~\ref{section_general_form_CSS}).

\section{The \CSS Algorithm} 
\label{sec:algorithms}
We now present the \CSS algorithm for combinatorial structures and describe its two components: a model tailored to combinatorial domains in Sect.~\ref{sec:statistical_model} and its acquisition function in Sect.~\ref{sec:acquisition_fn}.  Sect.~\ref{sec:css_alg} summarizes the algorithm and Sect.~\ref{section_CSS_SA} presents the variant~\CSSSA.
The time complexity is analyzed in Sect.~\ref{section_timecomplexity}.

\subsection{Statistical Model} 
\label{sec:statistical_model}
When developing a generative model for an expensive function $f(x): \domain \rightarrow \mathbb{R}$ defined on a combinatorial domain, it seems essential to model the interplay of elements. 
For example, in the above bike sharing application, the utility of placing a station at some location depends critically on the presence of other stations.
Similarly, the absorption of a medical drug depends on the combination of functional groups in the molecule.
A general model for~$f$ is thus given by
$\sum_{S \in 2^\domain} \alpha_S \prod_{i \in S}x_i$, where~$2^\domain$ is the power set of the domain and~$\alpha_S$ is a real-valued coefficient.
Clearly, this model is impractical due to the exponential number of monomials.
Thus, we consider restricted models that contain monomials up to order~$k$. 
A higher order increases the expressiveness of the model but also decreases the accuracy of the predictions when data is limited (e.g., see Ch.~14.6 in \citet{gelman_2014}).
We found that second-order models provide an excellent trade-off in practice (cp.\ Sect.~\ref{section_experiments} and~\ref{section_experiments_higher_order_models}). 
Thus, under our model, $x$ has objective value
\begin{equation} \label{eq:stat_model}
f_{\balpha}(x) = \alpha_0 + \sum_{j} \alpha_{j} x_{j} + \sum_{i,j > i} \alpha_{ij} x_{i} x_{j}.
\end{equation}
While the so-called \emph{interaction terms} are quadratic in $x \in \mathcal{D}$, the regression model is linear in $\balpha = (\alpha_{i},\alpha_{ij}) \in \mathbb{R}^{p}$ with~$p = 1 + d + \binom{d}{2}$.

\paragraph*{Sparse Bayesian Linear Regression:} \label{sec:sparse_blr}

To quantify the uncertainty in the model, we propose a Bayesian treatment for~$\balpha$.
For observations $(x^{(i)},y^{(i)}(x^{(i)}))$ with $i=1,\dots,N$, let~$\mathbf{X} \in \{0,1\}^{N \times p}$ be the matrix of predictors and $\mathbf{y} \in \mathbb{R}^{N}$ the vector of corresponding observations of $f$.
Using the data model, $y^{(i)}(x^{(i)}) = f(x^{(i)}) + \varepsilon^{(i)}$ where $\varepsilon^{(i)} \sim \mathcal{N}(0,\sigma^2)$, we have 
$\mathbf{y} \mid{} \mathbf{X},\boldsymbol{\alpha},\sigma^2 \sim \mathcal{N}(\mathbf{X}\boldsymbol{\alpha},\sigma^2I_{N})$.

One drawback of using a second-order model is that it has~$\Theta(d^2)$ regression coefficients which may result in high-variance estimators for the coefficients if data is scarce. 
To assert a good performance even for high-dimensional problems with expensive evaluations, we employ a \emph{sparsity-inducing prior}. 
We use the heavy-tailed horseshoe prior~\cite{carvalho_2010}:
\begin{align*}
\alpha_{k} \mid{} \beta_{k}^2,\tau^2,\sigma^2 &\sim \mathcal{N}(0,\beta_{k}^2\tau^2\sigma^2) \;\;\; k =1,\dots,p \\
\tau,\beta_{k} &\sim \mathcal{C}^{+}(0,1) \;\;\; k =1,\dots,p\\
P(\sigma^2) &= \sigma^{-2},
\end{align*}
where $\mathcal{C}^{+}(0,1)$ is the standard half-Cauchy distribution. In this model, the global, $\tau$, and the local, $\beta_{k}$, hyper-parameters individually shrink the magnitude of each regression coefficient.
Following \citet{makalic_2016}, we introduce the auxiliary variables~$\nu$ and~$\xi$ to re-parameterize the half-Cauchy densities using inverse-gamma distributions.
Then the conditional posterior distributions for the parameters are given by
\begin{align}
\label{eq:bayesian_posterior}
\boldsymbol{\alpha}|\cdot &\sim \mathcal{N}(\mathbf{A}^{-1}\mathbf{X}^{T}\mathbf{y},\sigma^2\mathbf{A}^{-1}), \\
\mathbf{A} &= (\mathbf{X}^{T}\mathbf{X} + \Sigma_{*}^{-1}), \Sigma_{*} = \tau^2\text{diag}(\beta_{1}^2,\dots,\beta_{p}^2) \nonumber \\
\sigma^2|\cdot &\sim IG \left(\frac{N+p}{2},\frac{(\mathbf{y} - \mathbf{X}\boldsymbol{\alpha})^{T}(\mathbf{y} - \mathbf{X}\boldsymbol{\alpha})}{2} {+} \frac{\boldsymbol{\alpha}^{T} \Sigma_{*}^{-1} \boldsymbol{\alpha}}{2} \right) \nonumber \\
\beta_{k}^2|\cdot &\sim IG \left(1, \frac{1}{\nu_{k}} + \frac{\alpha_{k}^2}{2\tau^2\sigma^2} \right) \;\;\; k =1,\dots,p \nonumber \\
\tau^2|\cdot &\sim IG\left(\frac{p+1}{2},\frac{1}{\xi} + \frac{1}{2\sigma^2} \sum_{k=1}^{p} \frac{\alpha_{k}^2}{\beta_{k}^2} \right) \nonumber \\
\nu_{k}|\cdot &\sim IG \left(1,1 + \frac{1}{\beta_{k}^2} \right) \;\;\; k =1,\dots,p \nonumber \\
\xi|\cdot &\sim IG \left(1,1 + \frac{1}{\tau^2} \right). \nonumber
\end{align}
Given these closed-form conditionals, we employ a Gibbs sampler to efficiently sample from the posterior over~$\balpha$. 
The complexity of sampling~$\boldsymbol{\alpha}$ is dominated by the cost of sampling from the multivariate Gaussian. This step has cost~$\mathcal{O}(p^3)$ for a na\"{i}ve implementation and hence can be prohibitive for a large number of predictors.
Instead we use the exact sampling algorithm of~\citet{bhattacharya_2016} whose complexity is~$\mathcal{O}(N^2p)$ and therefore nearly linear in $p$ whenever~$N \ll p$.
We have evaluated the different approaches and found that the proposed sparse regression performs well for several problems (see Sect.~\ref{section_validation_models} in the supplement for details).

We note that if the statistical model for $f$ is based on a maximum likelihood estimate (MLE) for $\balpha$ (see Sect.~\ref{section_mle_model}), the algorithm would exhibit a purely exploitative behavior and produce sub-optimal solutions.
Thus, it seems essential to account for the uncertainty in the model for the objective, which is accomplished by sampling the model parameters from the posterior over $\balpha$ and $\sigma^2$.

\subsection{Acquisition Function} 
\label{sec:acquisition_fn}
The role of the acquisition function is to select the next sample point in every iteration.
Ours is inspired by Thompson sampling \cite{thompson1933likelihood,thompson1935theory} (also see the excellent survey of~\citet{russo2017tutorial}) that samples a point~$x$ with probability proportional to~$x$ being an optimizer of the unknown function.
We proceed as follows.
Keeping in mind that our belief on the objective~$f$ at any iteration is given by the posterior on~$\balpha$, we sample~$\balpha_{t} \sim P(\balpha \mid \textbf{X}, \textbf{y})$ and want to find an~$\argmax_{x \in \domain} f_{\balpha_{t}}(x)$.
Since applications often impose some form of regularization on~$x$, we restate the problem as $\argmax_{x \in \mathcal{D}} f_{\balpha}(x) - \lambda \mathcal{P}(x)$, where~$\mathcal{P}(x) = \|x\|_1$ or~$\mathcal{P}(x) = \|x\|_2^2$ and thus cheap to evaluate.
Then, for a given~$\balpha$ and $\mathcal{P}(x) = \|x\|_1$, the problem is to obtain an
\begin{align}
& \argmax_{x \in \mathcal{D}} f_{\balpha}(x) - \lambda \mathcal{P}(x) \\
=& \argmax_{x \in \mathcal{D}} \sum_{j} (\alpha_{j} - \lambda) x_{j} + \sum_{i,j>i} \alpha_{ij} x_{i} x_{j}, \label{eq:acquisition_opt}
\end{align}
where~$x \in \{0,1\}^d$. Similarly, if $\mathcal{P}(x) = \|x\|_2^2$, the problem becomes
$\argmax_{x \in \mathcal{D}} \sum_{j} \alpha_{j} x_{j} + \sum_{i,j > i} (\alpha_{ij} - \lambda \delta_{ij}) x_{i}x_{j}$.
Thus, in both cases we are to solve a binary quadratic program of the form 
\begin{equation}
\label{Eq_binary_QP} 
\argmax_{x \in \mathcal{D}} x^T A x + b^T x,
\end{equation}
where~$\mathcal{D} = \{0,1\}^d$.
That is, we are to optimize a quadratic form over the vertices of the $d$-dimensional hypercube~${\cal H}_d$. 
This problem is known to be notoriously hard, not admitting exact solutions in polynomial time unless~$\P = \NP$~\cite{garey2002computers,charikar_2004}.

\paragraph*{Outline:}
We will show how to obtain an approximation efficiently.
First, we relax the quadratic program into a vector program, replacing the binary variables by high-dimensional vector-valued variables on the~$(d+1)$-dimensional unit sphere~${\cal S}_{d}$. 
Note that the optimum value attainable for this convex relaxation is at least as large as the optimum of Eq.~(\ref{Eq_binary_QP}).
We then rewrite this vector program as a semidefinite program (SDP) that can be approximated in polynomial time to a desired precision \cite{steurer2010fast,arora2016combinatorial,boyd2004convex}. 
The solution to the SDP is converted back into a collection of vectors. Finally, we apply the randomized rounding method of \citet{charikar_2004} to obtain a solution in~\domain.
We found that this procedure often produces an~$x^{(t)}$ that is (near)-optimal. Moreover, it has a robustness guarantee in the sense that the approximation error never deviates more than~$\mathcal{O}(\log d)$ from the optimum. 
(This bound requires that~$\alpha_0$ does not carry a negative weight that is large in absolute value compared to the optimal value of the~SDP, see \cite{charikar_2004} for details.)
Note that the worst case guarantee is essentially the best possible under standard complexity assumptions \cite{arora2005non}.

To convert the input domain to $\{-1,1\}^{d}$, we replace each variable~$x_i$ by $y_i = 2x_i - 1$ and accordingly adapt the coefficients by defining $\tilde{A} = A/4$, $c = b/4 + A^{T}\mathbf{1}/4$, and
$$B = \begin{bmatrix} \tilde{A} & c \\ c^{T} & 0 \end{bmatrix}.$$
Augmenting the state with an additional variable $y_{0}$, we can rewrite~\eqref{Eq_binary_QP} as the quadratic program $\argmax_{z} \; z^{T} B z$
with $z {=} [y,\,y_{0}] \in \{-1,1\}^{d+1}$. 
Thus, replacing~$y_i$ by the real-valued vector-variable~$\nu_i \in {\cal S}_{d}$ for~$0 {\leq} i {\leq} d$ we obtain the relaxation $\argmax \sum_{i,j} B_{i,j} \langle \nu_i,\nu_j\rangle $. This vector program is equivalent to the~SDP
\begin{equation}
\label{eq:sdp_relaxation}
\argmax_{Z \succeq 0} \; \text{Tr}(B^{T}Z) \quad \text{s.t.} \; \text{diag}(Z) = \text{I}_{d+1},
\end{equation}
where~$Z$ is a symmetric $(d{+}1) {\times} (d{+}1)$ real matrix. 

First we obtain a solution~$Z^{*}$ to~\eqref{eq:sdp_relaxation} that is then (approximately) factorized as $Z^{*} = (V^{*})^{T}V^{*}$, where $V^{*} \in \mathbb{R}^{(d+1) \times (d+1)}$ contains column vectors that satisfy the constraints $\| V^{*}_{i} \|_{2} = 1$ (i.e., $V^{*}_{i} \in \mathcal{S}_{d}$). 
Then, drawing a random vector $r \in \mathbb{R}^{d+1}$ with independent standard Gaussian entries, we apply the randomized geometric rounding procedure of \citet{charikar_2004} to obtain an approximate solution $y^{*} \in \{-1,1\}^{d}$ to the original quadratic program. 
Lastly, we apply the inverse transformation $x_i^{*}=(y_i^{*}+1)/2$ to recover a solution on the $d$-dimensional hypercube.

\subsection{Summary of the \CSS Algorithm} 
\label{sec:css_alg}
We now summarize the \CSS algorithm. Using an initial dataset of $N_{0}$ samples, it first computes the posterior on $f$ based on the sparsity-inducing prior. %

In the optimization phase, \CSS proceeds in iterations until the sample budget~$N_{max}$ is exhausted. 
In iteration $t =1,2,\ldots$, it samples the vector~$\balpha_{t}$ from the posterior over the regression coefficients that is defined by the parameters in Eq.~(\ref{eq:bayesian_posterior}). 
Now \CSS computes an approximate solution~$x^{(t)}$ for $\max_{x \in \{0,1\}^d} f_{\balpha_{t}} - \lambda \mathcal{P}(x)$ as follows:
first it transforms the quadratic model into an SDP, thereby relaxing the variables into vector-valued variables on the $(d+1)$-dimensional unit-sphere. 
This SDP is solved (with a pre-described precision) and the next point $x^{(t)}$ is obtained by rounding the vector-valued SDP solution.
The iteration ends after the posterior is updated with the new observation~$y^{(t)}$ at~$x^{(t)}$.
\CSS is summarized as Algorithm~\ref{alg:comb_structure_search}.
\begin{algorithm}[!ht]
\caption{Bayesian Optimization of Combinatorial Structures}
\label{alg:comb_structure_search}
\begin{algorithmic}[1]
\STATE {\bfseries Input:} Objective function $f(x) - \lambda\mathcal{P}(x)$; Sample budget~$N_{max}$; Size of initial dataset~$N_0$.
\STATE Sample initial dataset~$D_0$.
\STATE Compute the posterior on~$\balpha$ given the prior and~$D_0$.
\FOR{$t = 1$ {\bfseries to} $N_{max} - N_0$}
\STATE Sample coefficients $\balpha_{t} \sim P(\balpha \mid \textbf{X}, \textbf{y})$.
\STATE Find approximate solution $x^{(t)}$ for $\max_{x \in \mathcal{D}} f_{\balpha_{t}}(x) - \lambda \mathcal{P}(x)$.
\STATE Evaluate $f(x^{(t)})$ and append the observation~$y^{(t)}$ to~$\mathbf{y}$.
\STATE Update the posterior $P(\balpha \mid \textbf{X}, \textbf{y})$.
\ENDFOR
\STATE \textbf{return} $\argmax_{x \in \mathcal{D}} f_{\balpha_{t}}(x) - \lambda \mathcal{P}(x)$.
\end{algorithmic}
\end{algorithm}

\subsection{\CSSSA: A Low-Complexity Variant of~\CSS}
\label{section_CSS_SA}
We propose a variant of~\CSS that replaces semidefinite programming by stochastic local search.
In our experimental evaluation, the solver takes only a few seconds to obtain a solution to the semidefinite program and will scale easily to a few hundred dimensions.  
While semidefinite programs can be approximated to a given precision in polynomial time, their complexity might become a bottleneck in future applications when the dimensionality grows large. 
Therefore, we also investigated alternative techniques to solve the problem in~(\ref{Eq_binary_QP}) and have found good performance with stochastic local search, specifically with simulated annealing.

Simulated annealing (\SA) performs a random walk on~\domain, starting from a point chosen uniformly at random. Let~$x^{(t)}$ be the point selected in iteration~$t$. Then the next point~$x^{(t+1)}$ is selected in the \emph{neighborhood} $N(x^{(t)})$ that contains all points with Hamming distance at most one from~$x^{(t)}$.
\SA picks $x \in N(x^{(t)})$ uniformly at random and evaluates~$\mathrm{obj}(x)$: 
If the observed objective value is better than the observation for~$x^{(t)}$, \SA sets $x^{(t+1)} = x$. 
Otherwise, the point is adopted with probability $\exp((\mathrm{obj}(x) - \mathrm{obj}(x^{(t)}))/T_{t+1})$, where~$T_{t+1}$ is the current temperature. 
\SA starts with a high $T$ that encourages exploration and cools down over time to zoom in on a good solution.

In what follows, \CSSSDP denotes the implementation of~\CSS that leverages semidefinite programming. The implementation that uses simulated annealing (\SA) is denoted by~$\CSSSA$.

\subsection{Time Complexity}
\label{section_timecomplexity}
Recall from Sect.~\ref{sec:statistical_model} that the computational cost of sampling from the posterior over~$\balpha$ and~$\sigma^2$ is bounded by~$\mathcal{O}(N^2 p)$, where~$p = \Theta(d^2)$ for the second-order model and~$N$ is the number of samples seen so far.
The acquisition function is asymptotically dominated by the cost of the SDP solver, which is polynomial in~$d$ for a given precision~$\varepsilon$.
Therefore, the total running time of a single iteration of~\CSSSDP is bounded by~$\mathcal{O}(N^2 d^2 + poly(d,\frac{1}{\varepsilon}))$.

A single iteration of~\CSSSA on the other hand has time complexity~$\mathcal{O}(N^2 d^2)$, since simulated annealing runs in~$\mathcal{O}(d^2)$ steps for the temperature schedule of \citet{spears_1993}.
We point out that the number of alternatives is exponential in~$d$, thus the running times of \CSSSDP and~\CSSSA are only \emph{logarithmic} in the size of the domain that we optimize over.

\section{Numerical Results} 
\label{sec:num_results}
\label{section_experiments}
We conduct experiments on the following benchmarks: (1) binary quadratic programming with $d=10$ variables (Sect.~\ref{sec:quadratic_test}), (2) sparsification of Ising models with~$d=24$ edges (Sect.~\ref{sec:sparse_ising}), (3) contamination control of a food supply chain with $d=25$ stages (Sect.~\ref{sec:cont_control}), (4) complexity reduction of an aero-structural multi-component system with $d=21$ coupling variables (Sect.~\ref{sec:aerospace_test}).
We evaluate the variants of the \CSS algorithm described in Sect.~\ref{sec:algorithms} and compare them to the following methods from machine learning and combinatorial optimization.

\emph{Expected improvement} (\EI) with a Gaussian process based model \cite{jsw98,snoek_2012} typically performs well for noise-free functions. \EI uses the popular one-hot encoding that coincides with the tailored kernel of \citet{hutter2009automated} for binary variables. 
Although the computational cost for selecting the next candidate point is relatively low compared to other acquisition functions, \EI is considerably more expensive than the other methods (see also Sect.~\ref{section_wallclock_times}).
\SMAC \cite{hutter2011sequential} addresses this problem by performing a local search for a candidate with high expected improvement. It uses a random forest-based model that is able to handle categorical and integer-valued variables.

\emph{Sequential Monte Carlo particle search} (\PS) \cite{schafer_2013} is an evolutionary algorithm that maintains a population of candidate solutions. \PS is robust to multi-modality and often outperforms local search and simulated annealing for combinatorial domains \cite{del2006sequential,schafer_2013}.

\emph{Simulated annealing} (\texttt{SA}) is known for its excellent performance on hard combinatorial problems~\cite{spears_1993,pankratov_2010,poloczek_2017}. 

Starting at a randomly chosen point, \emph{oblivious local search} (\OLS) \cite{kmsv98} evaluates in every iteration all points with Hamming distance one from its current point and adopts the best. We are interested in the search performance of \OLS relative to its sample complexity. At each iteration, \OLS requires~$d$ function evaluations to search within the neighborhood of the current solution.
We also compare to \emph{random search} (\RS) of~\citet{bb12}.

We report the function value returned after~$t$ evaluations, averaged over at least~100 runs of each algorithm for the first three problems. 
Intervals stated in tables and error bars in plots give the mean $\pm$ 2 standard errors. Bold entries in tables highlight the best mean performance for each choice of~$\lambda$. 
\CSS and \EI are given identical initial datasets in every replication. These datasets were drawn via Monte Carlo sampling. %
Algorithms that do not take an initial dataset are allowed an equal number of `free' steps before counting their function evaluations.
The implementations of the above algorithms and the variants of \CSS are available at \url{https://github.com/baptistar/BOCS}.

\subsection{Binary Quadratic Programming} 
\label{sec:quadratic_test}
\begin{figure}[!ht]
	\centering
	\includegraphics[width=0.9\linewidth]{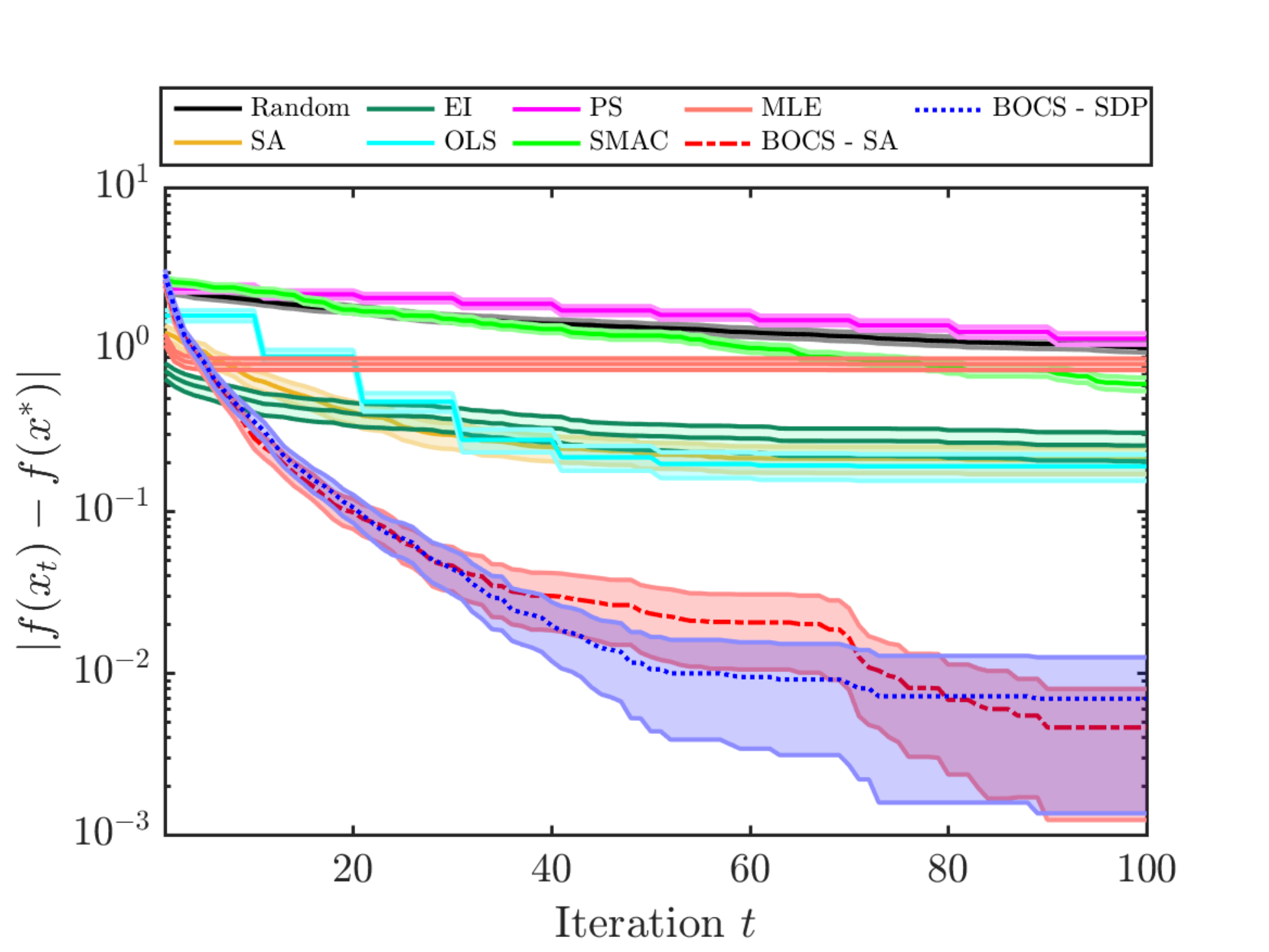}
	\vspace{-10pt}
	\caption{Random BQP instances with $L_{c} = 10$ and $\lambda = 0$: Both variants of \CSS outperform the competitors substantially.}
	\label{fig:quad_utility_gap_alpha1_lambda0}
\end{figure}

The objective in the binary quadratic programming problem (BQP) is to maximize a quadratic function with $\ell_{1}$ regularization, $f(x) - \lambda \mathcal{P}(x) = x^{T}Qx - \lambda \|x\|_1$, over~$\{0,1\}^d$. 
$Q \in \mathbb{R}^{d \times d}$ is a random matrix with independent standard Gaussian entries that is multiplied element-wise by a matrix $K \in \mathbb{R}^{d \times d}$ with entries $K_{ij} = \exp(-(i-j)^2/L_{c}^2)$. The entries of~$K$ decay smoothly away from the diagonal with a rate determined by the correlation length~$L_{c}^2$. 
We note that, as the correlation length increases, $Q$ changes from a nearly diagonal to a denser matrix, making the optimization more challenging.
We set~$d=10$, sampled~$50$ independent realizations for~$Q$, and ran every algorithm 10-times on each instance with different initial datasets. Bayesian optimization algorithms received identical initial datasets of size~$N_0 = 20$. 
Recall the performance at step~$t$ of the other algorithms (i.e., \SA, \OLS, and \RS) corresponds to the~$(t{+}N_0)$-th function evaluation.
For $\lambda = 0$ and $L_{c} = 10$, Fig.~\ref{fig:quad_utility_gap_alpha1_lambda0} reports the simple regret after step~$t$ , i.e., the absolute difference between the global optimum and the solution returned by the respective algorithm. 

We see that both variants of \CSS perform significantly better than the competitors. \CSSSDP and the variant \CSSSA based on stochastic local search are close with the best performance.
\EI and \SA make progress slowly, whereas the other methods are clearly distanced.
When considering the performance of \OLS, we note that a deterministic search over a $1$-flip neighborhood seems to make progress, but is eventually stuck in local optima.
Similarly, \MLE plateaus quickly. We discuss this phenomenon below.

\begin{figure}[!ht]
	\centering
	\includegraphics[width=0.9\linewidth]{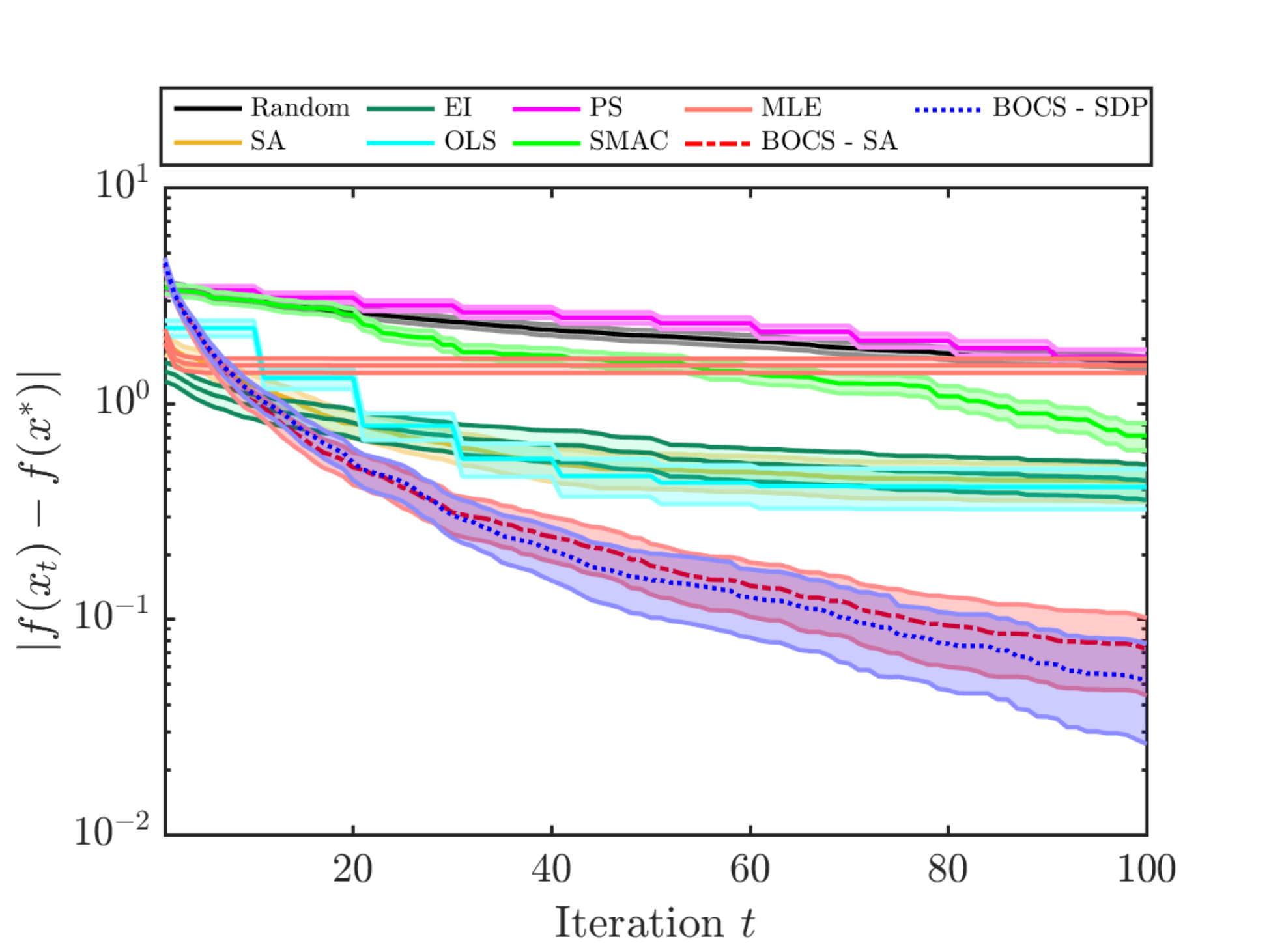}
	\vspace{-10pt}
	\caption{Random BQP instances with $L_{c} {=} 100$ and $\lambda {=} 1$: Both versions of \CSS outperform the other methods.}
	\label{fig:quad_utility_gap_alpha100_lambda1}
\end{figure}

We also studied the performance for $L_{c} {=} 100$ and~$\lambda{=}1$, see Fig.~\ref{fig:quad_utility_gap_alpha100_lambda1}.
Again, \CSSSDP performs substantially better than the other algorithms, followed by~\CSSSA. Table~\ref{results_quad_test} compares the performances of \EI and \CSS across other settings of~$L_c$ and~$\lambda$.
\begin{table}[!ht]
	\caption{The simple regret after 100 iterations for $10$-dimensional BQP instances. The entries have been multiplied by~$10$. The best performance for each setting is set in bold.}
	\label{results_quad_test}
	\vskip 0.05in
	\begin{center}
		\begin{small}
			\begin{sc}
				\begin{tabular}{lccc}
					\toprule
					$(L_{c},\lambda)$ & \EI & \CSSSA & \CSSSDP \\
					\midrule
					$(1,0)$   	    & $0.49 \pm 0.13$ & $\textbf{0.02} \pm 0.02$ & $0.03 \pm 0.02$ \\
					$(1,10^{-4})$   & $0.50 \pm 0.12$ & $\textbf{0.02} \pm 0.01$ & $0.03 \pm 0.03$ \\
					$(1,10^{-2})$   & $0.54 \pm 0.12$ & $\textbf{0.02} \pm 0.02$ & $0.05 \pm 0.05$ \\[0.1cm]
					
					$(10,0)$   		& $2.54 \pm 0.51$ & $\textbf{0.07} \pm 0.05$ & $0.07 \pm 0.05$ \\
					$(10,10^{-4})$  & $2.49 \pm 0.44$ & $\textbf{0.06} \pm 0.04$ & $0.08 \pm 0.05$ \\
					$(10,10^{-2})$  & $2.27 \pm 0.40$ & $\textbf{0.04} \pm 0.04$ & $0.10 \pm 0.06$ \\[0.1cm]
					
					$(100,0)$   	& $3.38 \pm 0.70$ & $0.15 \pm 0.07$ & $\textbf{0.11} \pm 0.06$ \\
					$(100,10^{-4})$ & $4.07 \pm 0.77$ & $0.16 \pm 0.08$ & $\textbf{0.15} \pm 0.08$ \\
					$(100,10^{-2})$ & $4.25 \pm 0.78$ & $0.17 \pm 0.09$ & $\textbf{0.13} \pm 0.07$ \\[0.1cm]
					\bottomrule
				\end{tabular}
			\end{sc}
		\end{small}
	\end{center}
	\vskip -0.1in
\end{table}
\texttt{MLE} is derived from~\CSSSA by setting the regression weights to a maximum likelihood estimate (see Sect.~\ref{section_mle_model}). 
We witnessed a purely exploitative behavior of this algorithm and an inferior performance that seems to plateau.
This underlines the importance of sampling from the posterior of the regression weights, which enables the algorithm to explore the model space, resulting in significantly better performance.

\subsection{Sparsification of Ising Models} 
\label{sec:sparse_ising}
We consider zero-field Ising models that admit a distribution $p(z) = \frac{1}{Z^{p}}\exp(z^{T}J^{p}z)$ for $z \in \{-1,1\}^{n}$, where $J^{p} \in \mathbb{R}^{n \times n}$ is a symmetric interaction matrix and $Z^{p}$ is the partition function. The support of the matrix $J^{p}$ is represented by a graph $\mathcal{G}^{p}$ = $([n],\mathcal{E}^{p})$ that satisfies $(i,j) \in \mathcal{E}^{p}$ if and only if $J^{p}_{ij} \neq 0$ holds.

Given $p(z)$, the objective is to find a close approximating distribution $q_{x}(z) = \frac{1}{Z^{q}}\exp(z^{T}J^{q}z)$ while minimizing the number of edges in $\mathcal{E}^{q}$. 
We introduce variables $x \in \{0,1\}^{|\mathcal{E}^{p}|}$ that indicate if each edge is present in $\mathcal{E}^{q}$ and set the edge weights as $J_{ij}^{q} = x_{ij}J_{ij}^{p}$.
The distance between $p(z)$ and $q_{x}(z)$ is measured by the Kullback-Leibler (KL) divergence
\begin{equation*} 
\label{eq:opt_ising}
D_{KL}(p||q_{x}) = \sum_{(i,j) \in \mathcal{E}^{p}} (J^{p}_{ij} - J^{q}_{ij}) \mathbb{E}_{p}[z_{i}z_{j}] + \log \left(\frac{Z_{q}}{Z_{p}} \right).
\end{equation*}

\begin{figure}
	\centering
	\includegraphics[width=0.9\linewidth]{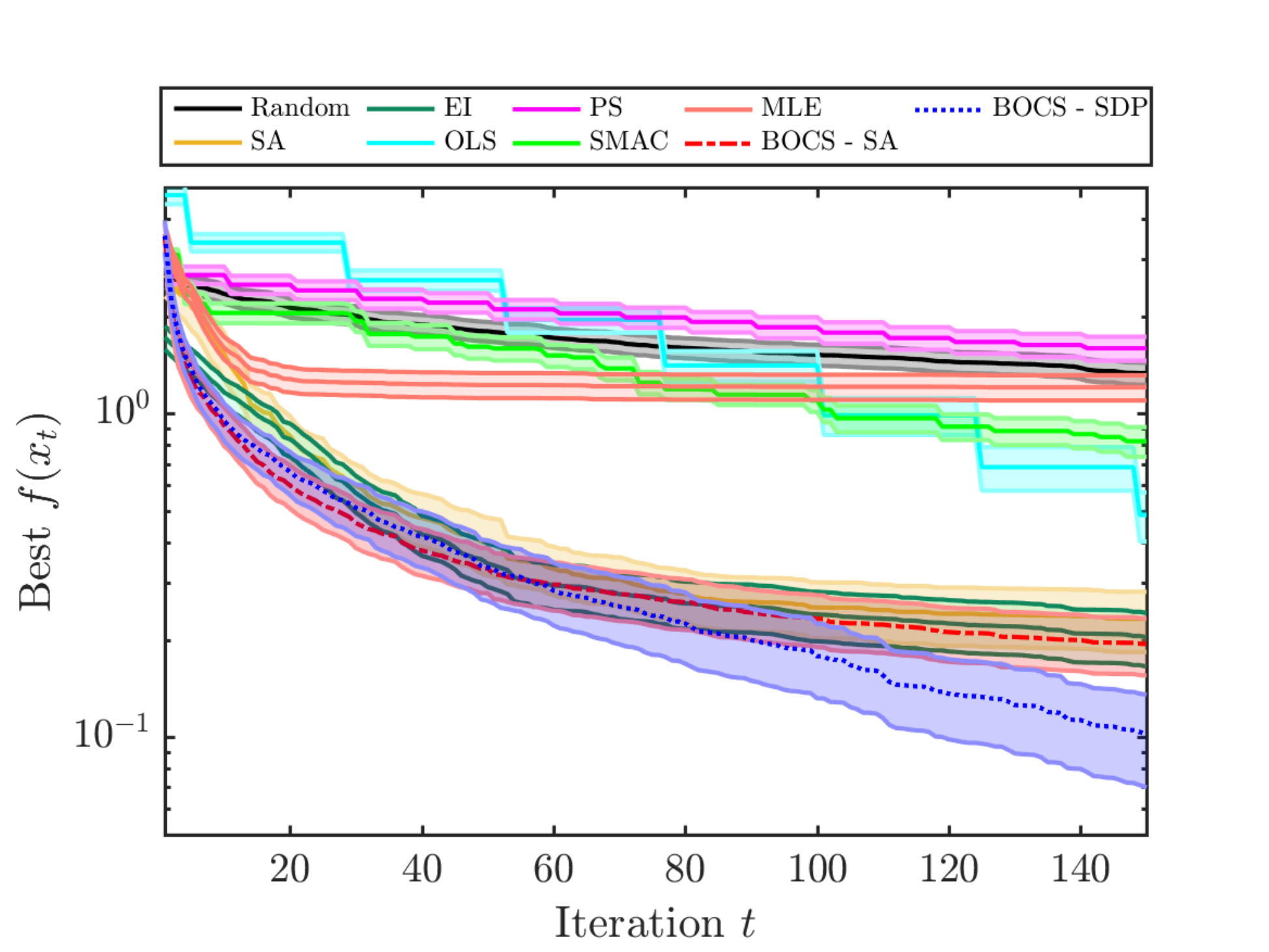}
	\vspace{-10pt}
	\caption{Sparsification of Ising models ($\lambda = 10^{-4}$): \CSSSDP performs best followed by~\CSSSA. \EI and \SA also show good performance. Due to the size of the search space and the evaluation cost of the objective, we report the average best function values after~$t$ iterations rather than the simple regret.}
	\label{fig:ising_best_soln_lambda0}
\end{figure}

Note that the cost of computing the ratio of the partition functions grows exponentially in~$n$, which makes the KL divergence an expensive-to-evaluate function. 
Summing up, the goal is to obtain an $\argmin_{x \in \{0,1\}^{d}} D_{KL}(p||q_{x}) + \lambda \|x\|_1$.
The experimental setup consists of $4 \times 4$ zero-field Ising models with grid graphs, i.e., $n = 16$ nodes and $d = 24$ edges. 
The edge parameters are chosen independently and uniformly at random in the interval~$[.05,5]$. The sign of each parameter is positive with probability~$1/2$ and negative otherwise.
The initial dataset contains~$N_{0} = 20$ points.
We note that with a cost of about~$1.8s$ for a single evaluation of the KL divergence, enumerating all $|\domain| = 2^{24}$ inputs to obtain an optimal solution is infeasible.
Thus, we report the best value obtained after iteration~$t$  rather than the simple regret. Fig.~\ref{fig:ising_best_soln_lambda0} depicts the mean performance with 95\% confidence intervals for $\lambda = 10^{-4}$, when averaged over $10$ randomly generated Ising models and $10$ initial data sets $D_{0}$ for each model.
The statistics for other values of $\lambda$ are stated in Table~\ref{tab:results_ising}. 
Initially, \CSSSDP, \CSSSA, \EI and \SA show a similar performance. As the number of samples increases, \CSSSDP obtains better solutions and in addition achieves a lower variability across different instances of Ising models. 

\begin{table}%
	\caption{Sparsification of Ising models: \CSSSDP obtains the best function values for all three settings of~$\lambda$, here measured after 150 iterations. We also note that the \CSSSDP algorithm has the lowest variability over 10 random Ising models. \label{tab:results_ising}}
	\vskip 0.05in
	\begin{center}
		\begin{small}
			\begin{sc}
				\begin{tabular}{lccc}
					\toprule
					$\lambda$   & \SA & \EI & \OLS \\
					\midrule
					$0$   	    & $0.21 \pm 0.05$ & $0.20 \pm 0.04$ & $0.54 \pm 0.09$ \\
					$10^{-4}$   & $0.23 \pm 0.05$ & $0.20 \pm 0.04$ & $0.49 \pm 0.08$ \\
					$10^{-2}$   & $0.39 \pm 0.05$ & $0.39 \pm 0.04$ & $0.67 \pm 0.09$ \\[0.1cm]
					
					\toprule
					$\lambda$   & \MLE-\SA & \CSSSA & \CSSSDP \\
					\midrule
					$0$   	    & $1.19 \pm 0.11$ & $0.19 \pm 0.04$ & $\textbf{0.11} \pm 0.04$ \\
					$10^{-4}$   & $1.21 \pm 0.11$ & $0.19 \pm 0.04$ & $\textbf{0.10} \pm 0.03$ \\
					$10^{-2}$   & $1.37 \pm 0.11$ & $0.37 \pm 0.04$ & $\textbf{0.33} \pm 0.04$ \\[0.1cm]
					
					\bottomrule
				\end{tabular}
			\end{sc}
		\end{small}
	\end{center}
	\vskip -0.1in
\end{table}

\subsection{Contamination Control} 
\label{sec:cont_control}
The contamination control problem~\cite{hu_2010} considers a food supply with~$d$ stages that may be contaminated with pathogenic microorganisms. Specifically, we let random variable~$Z_{i}$ denote the fraction of contaminated food at stage~$i$ for $1 {\leq} i {\leq }d$. 
$(Z)$ evolves according to a random process that we describe next. 
At each stage~$i$, a prevention effort can be made to decrease the contamination by a random rate~$\Gamma_{i}$, incurring a cost~$c_{i}$.
If no prevention effort is taken, the contamination spreads with rate given by random variable~$\Lambda_{i}$.
This results in the recursive equation $Z_{i} = \Lambda_{i}(1 - x_{i})(1 - Z_{i-1}) + (1 - \Gamma_{i}x_{i})Z_{i-1}$, where~$x_{i} \in \{0,1\}$ is the decision variable associated with the prevention effort at stage~$i$.
Thus, the goal is to decide for each stage whether to implement a prevention effort in order to minimize the cost while ensuring the fraction of contaminated food does not exceed an upper limit~$U_{i}$ with probability at least~$1 - \epsilon$. 
The random variables $\Lambda_{i}, \Gamma_{i}$ and the initial contamination fraction~$Z_1$ follow beta-distributions, whereas $U_{i} {=} 0.1$ and~$\epsilon = 0.05$.

We consider the Lagrangian relaxation of the problem that is given by
\begin{equation} \label{eq:obj_foodcontrol}
\argmin_{x} \sum_{i=1}^{d} \left[c_{i}x_{i} + \frac{\rho}{T} \sum_{k=1}^{T} {1}_{\{Z_{k} > U_{i} \}} \right] + \lambda \|x\|_1,
\end{equation}
where each violation is penalized by~$\rho = 1$.
Recall that we have~$d=25$ stages.
We set $T = 10^{2}$, hence the objective requires $T$ simulations of the random process and is expensive to evaluate. 
The $\ell_{1}$ regularization term encourages the prevention efforts to occur at a small number of stages. 

\begin{figure}[ht]
	\centering
	\includegraphics[width=0.9\linewidth]{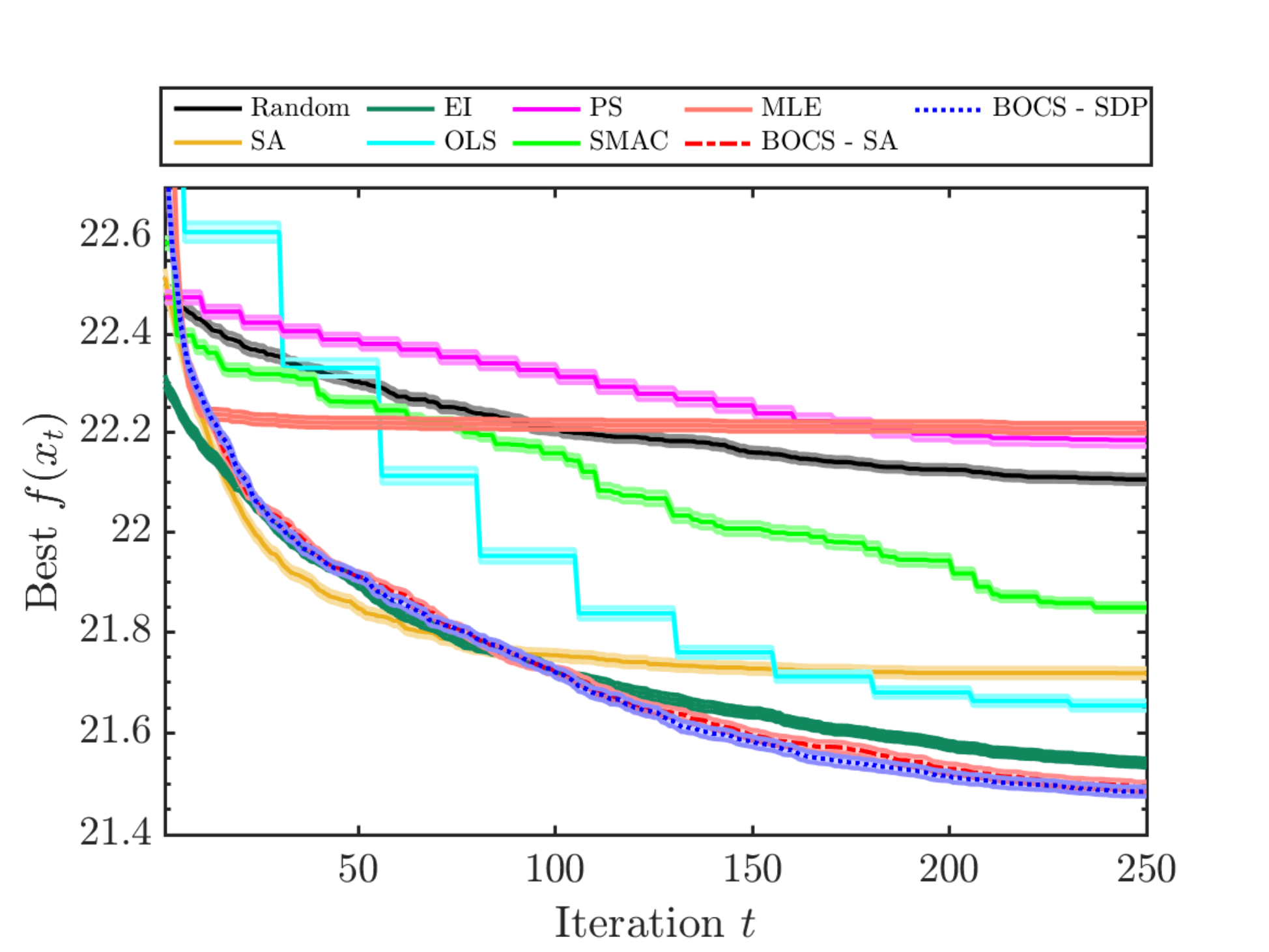}
	\vspace{-10pt}
	\caption{Contamination control ($\lambda = 10^{-2})$: Initially \SA performs best but then is trapped in a local optimum. As the number of samples increases, \CSSSDP obtains the best contamination prevention schedules, followed by \CSSSA and \EI.}
	\label{fig:foodcontrol_best_soln_lambda1em2}
\end{figure}

The mean objective value (with 95\% confidence intervals) of the best solution found after~$t$ iterations is shown in Fig.~\ref{fig:foodcontrol_best_soln_lambda1em2} for $\lambda = 10^{-2}$. 
Table~\ref{tab:results_foodcontrol} compares the performances for other values of $\lambda$ after 250 iterations.
\CSSSDP achieves the best performance in all scenarios.

\begin{table}[!ht]
	\caption{Contamination control: \CSSSA and \CSSSDP obtain the best function values for all three settings of~$\lambda$, here measured after 250 iterations. \label{tab:results_foodcontrol}}
	\vskip 0.05in
	\begin{center}
		\begin{small}
			\begin{sc}
				\begin{tabular}{lccc}
					\toprule
					$\lambda$   & \SA & \EI & \OLS \\
					\midrule
					$0$   	    & $21.58 \pm 0.01$ & $21.39 \pm 0.01$ & $21.54 \pm 0.01$ \\
					$10^{-4}$   & $21.60 \pm 0.01$ & $21.40 \pm 0.01$ & $21.51 \pm 0.01$ \\
					$10^{-2}$   & $21.72 \pm 0.01$ & $21.54 \pm 0.01$ & $21.65 \pm 0.01$ \\
					$1$    		& $\textbf{23.33} \pm 0.01$ & $24.71 \pm 0.02$ & $25.12 \pm 0.08$ \\[0.1cm]
					
					\toprule
					$\lambda$   & \MLE-\SA & \CSSSA & \CSSSDP \\
					
					\midrule
					$0$   	    & $22.02 \pm 0.01$ & $21.35 \pm 0.01$ & $\textbf{21.34} \pm 0.01$ \\
					$10^{-4}$   & $22.03 \pm 0.01$ & $21.36 \pm 0.01$ & $\textbf{21.35} \pm 0.01$ \\
					$10^{-2}$   & $22.21 \pm 0.02$ & $21.49 \pm 0.01$ & $\textbf{21.48} \pm 0.01$ \\
					$1$   	    & $\textbf{23.33} \pm 0.01$ & $\textbf{23.33} \pm 0.01$ & $\textbf{23.33} \pm 0.01$ \\[0.1cm]
					
					\bottomrule
				\end{tabular}
			\end{sc}
		\end{small}
	\end{center}
	\vskip -0.1in
\end{table}
While \SA and \EI initially perform well (see Fig.~\ref{fig:foodcontrol_best_soln_lambda1em2}),  \SA becomes stuck in a local optimum. 
After about 80 iterations, \CSS typically finds the best prevention control schedules, and improves upon the solution found by \EI.

\subsection{Aero-structural Multi-Component Problem} \label{sec:aerospace_test}
We study the aero-structural problem of~\citet{jasa_2017} that is composed of two main components, aerodynamics and structures (see Fig.~\ref{fig:aerostruct_ref_model}).
These blocks are coupled by~$21$ coupling variables that describe how aerodynamic properties affect the structure and vice versa, how loads and deflections affect the aerodynamics. 
\tikzset{
	dis_block_as/.style   = {draw, fill=blue!20, thick, rectangle, rounded corners, text centered, text width=2.0cm, minimum height = 1cm},
	inout_block_eng/.style = {text centered},
	coupling_node/.style   = {midway, text centered, text width=1.1cm}
}
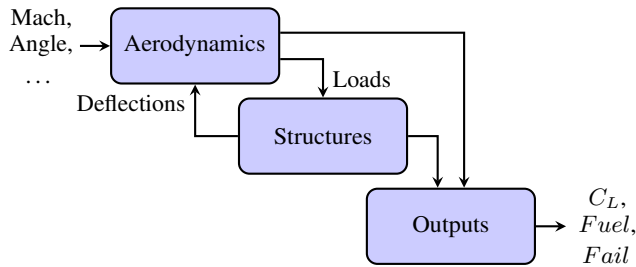
\begin{figure}
	\centering
	\begin{tikzpicture}[auto, thick, node distance=1.5cm]
	
	\draw

	node at (0,0)  	 	[dis_block_as] (aero) {\footnotesize Aerodynamics}
	node at (1.7,-1.2)  [dis_block_as] (str) {\footnotesize Structures}
	node at (3.4,-2.4)	[dis_block_as] (out) {\footnotesize Outputs}

	node [inout_block_eng, left=0.4cm of aero, text width=0.8cm] (inputs) {\small Mach,\\ Angle,\\ \dots}
	node [inout_block_eng, right=0.4cm of out, text width=0.8cm] (outputs) {\small $C_{L}$,\\ $Fuel$,\\ $Fail$};

	\draw[->, >=stealth] ([yshift=-0.5em]aero.east)  -| (str.north) node [above right] {\small Loads};
	\draw[->, >=stealth] (str.west) -| (aero.south) node [below left] {\small Deflections};
	\draw[->, >=stealth] ([yshift=+0.5em]aero.east)  -| ([xshift=+0.5em]out.north);
	\draw[->, >=stealth] (str.east)  -| ([xshift=-0.5em]out.north);

	\draw[->, >=stealth] (out.east) -- (outputs);
	\draw[->, >=stealth] (inputs.east)  -- (aero.west);
	
	\end{tikzpicture}
	\vspace{-20pt}
	\caption{The aero-structural model of~\citet{jasa_2017}: The arrows indicate the flow of information between components. Note that the loop requires a fixed point solve whose computational cost increases quickly with the number of involved coupling variables.}
	\label{fig:aerostruct_ref_model}
\end{figure}
For a set of inputs, e.g., airspeed, angle of the airfoil etc.,  with prescribed (Gaussian) probability distributions, the model computes the uncertainty in the output variables $\mathbf{y}$, which include the lift coefficient~$C_{L}$, the aircraft's fuelburn~$Fuel$, and a structural stress failure criteria~$Fail$. However, the amount of coupling in the model contributes significantly to the computational cost of performing this uncertainty quantification (UQ).

\begin{figure}[!h]
	\centering
	\includegraphics[width=0.9\linewidth]{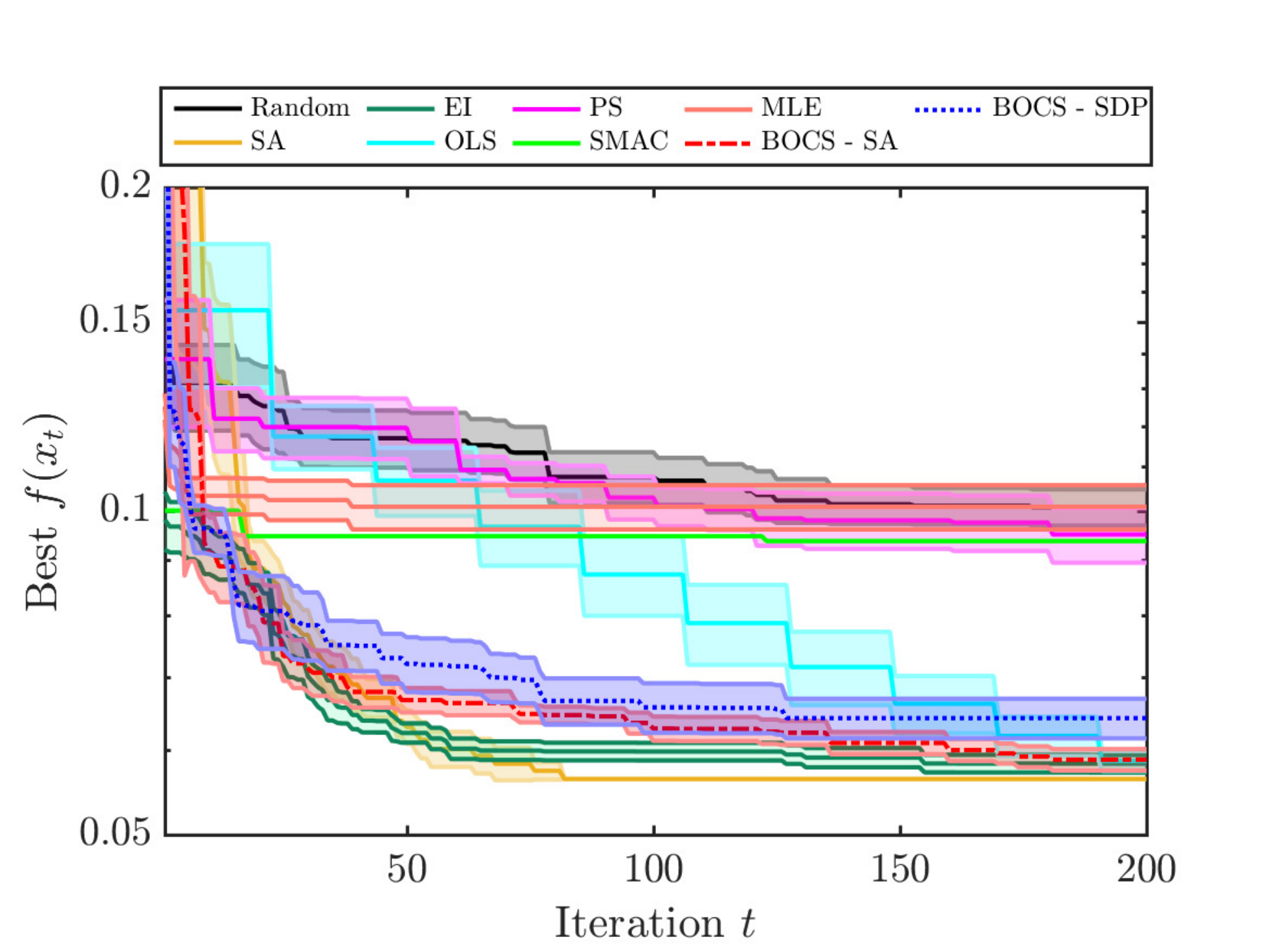}
	\vspace{-10pt}
	\caption{The aero-structural multi-component problem ($\lambda {=} 10^{-2}$): \SA performs best followed by \EI, \CSSSDP, and \CSSSA.}
	\label{fig:aerostruct_best_soln_lambda1em2}
\end{figure}

To accelerate the UQ process, we wish to identify a model with a reduced number of coupling variables that accurately captures the probability distribution of the output variables, $\pi_{\mathbf{y}}$. 
Let~$x \in \{0,1\}^{d}$ represent the set of `active' coupling variables: $x_{i} = 0$ denotes that coupling~$i$ from the output of one discipline is ignored and its input to another discipline is fixed to a prescribed value.
The effect of this perturbation on the model outputs is measured by the KL divergence between~$\pi_{\mathbf{y}}$, i.e., the output distribution of the reference model, and the output distribution~$\pi_{\mathbf{y}}^{x}$ for the model with coupling variables~$x$. 
Thus, the problem is to find an $\argmin_{x \in \{0,1\}^d} D_{KL}(\pi_{\mathbf{y}} || \pi^{x}_{\mathbf{y}}) + \lambda \|x\|_1$,
where~$D_{KL}(\pi_{\mathbf{y}} || \pi^{x}_{\mathbf{y}})$ is expensive to evaluate and~$\lambda \geq 0$ trades off accuracy and sparsity of the model. 
Fig.~\ref{fig:aerostruct_best_soln_lambda1em2} shows the average performances for~$\lambda = 10^{-2}$. \SA has the best overall performance, followed closely by \EI, \CSSSDP, and \CSSSA that have a similar convergence profile.

\begin{figure}[!h]
	\centering
	\includegraphics[width=0.45\linewidth]{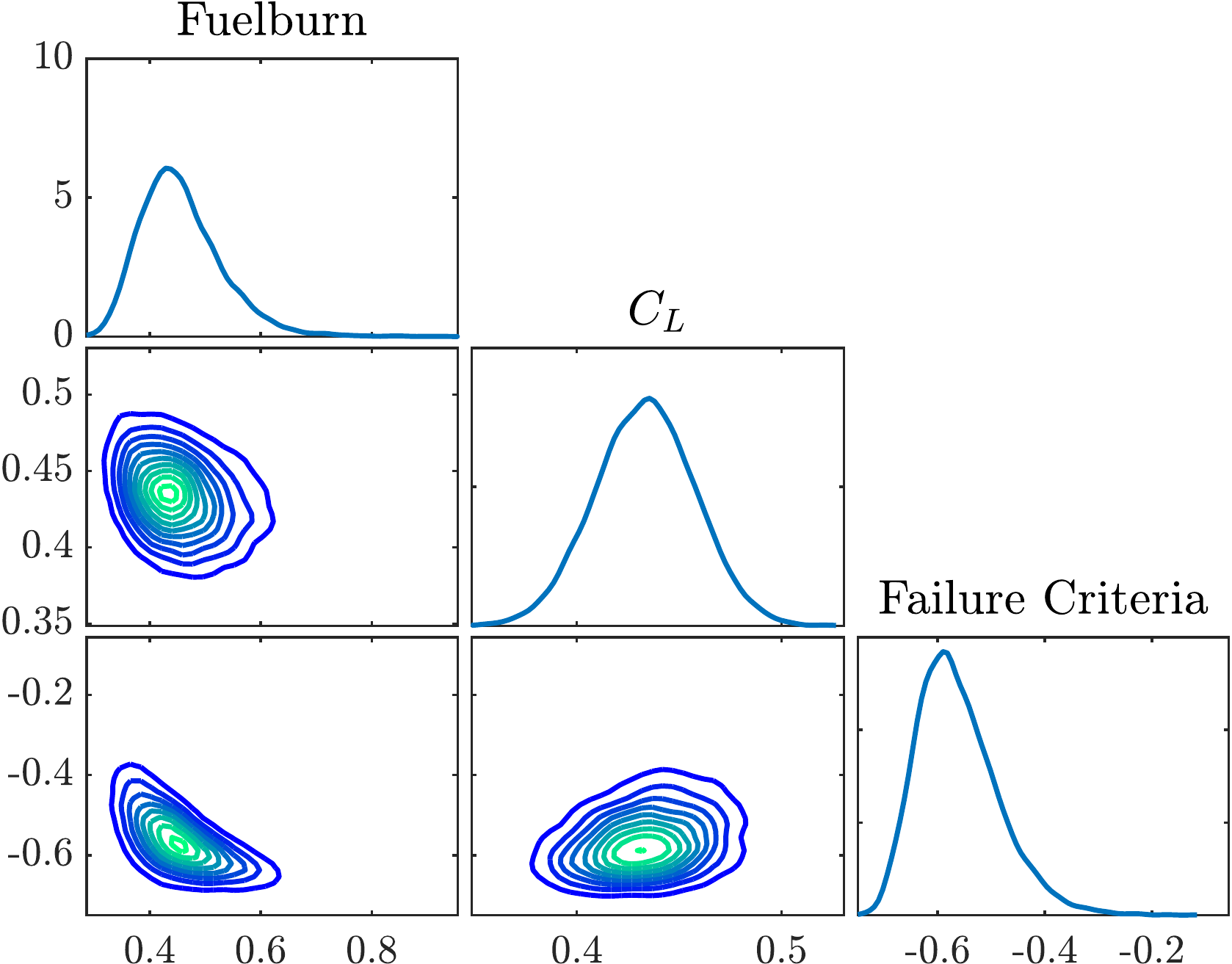}
	\hspace{5pt}
	\includegraphics[width=0.45\linewidth]{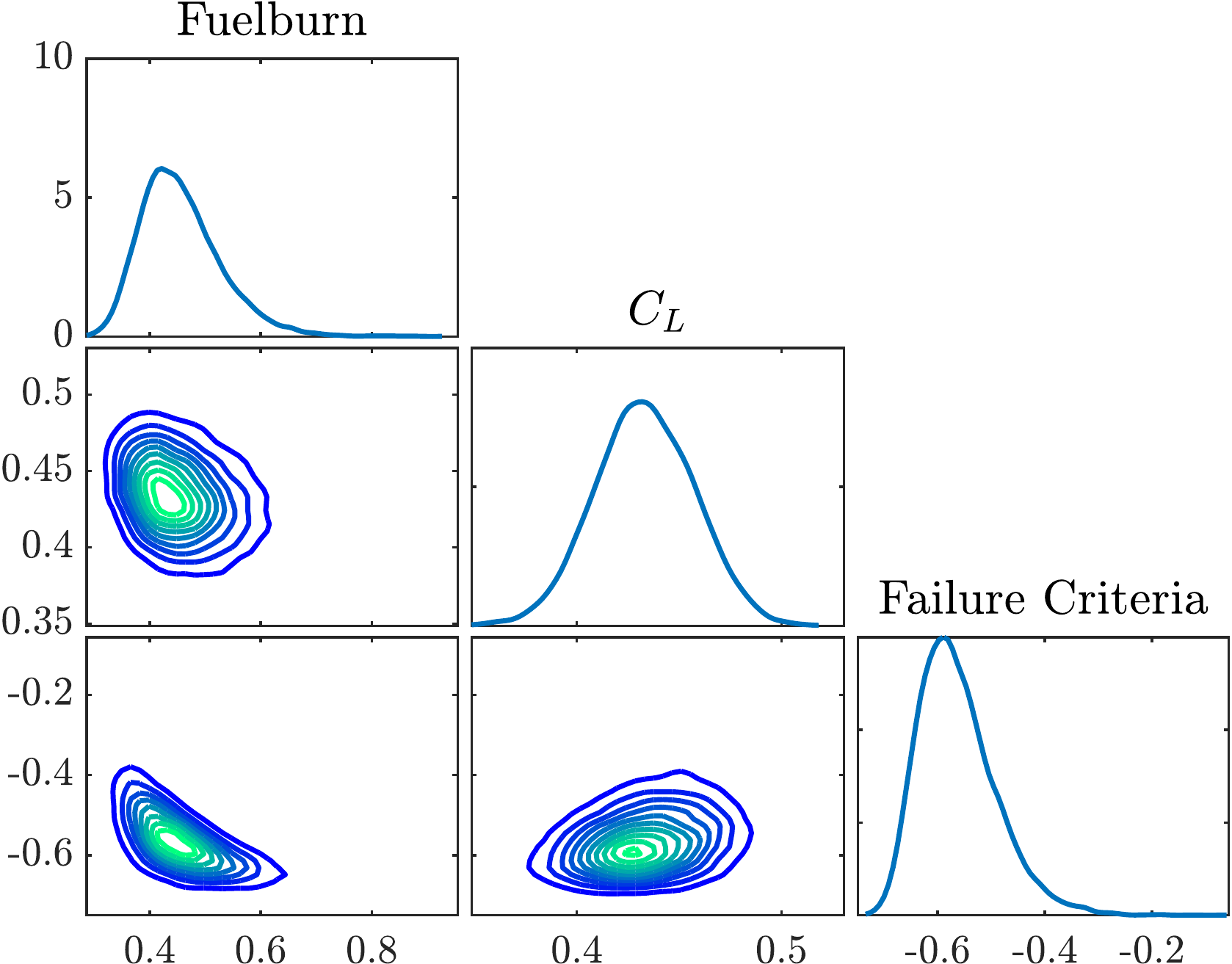}
	\vspace{-10pt}
	\caption{The aero-structural problem: the univariate and bivariate marginals of the output distribution for the reference (left) and a sparsified model (right) found by \CSS for $\lambda=10^{-2}$.
	\label{fig:aerostruct_soln_dist}}
\end{figure}

Fig.~\ref{fig:aerostruct_soln_dist} shows the output distribution of the reference model $\pi_{\mathbf{y}}$ (left), and the output distribution of a sparsified model $\pi_{\mathbf{y}}^{x}$ found by \CSS for $\lambda = 10^{-2}$ (right). We note that the distribution is closely preserved, while the sparsified solution only retains 5 out of the 21 active coupling variables present in the reference model.

\section{Conclusion}
\label{section_conclusion}
We have proposed the first algorithm to optimize expensive-to-evaluate black-box functions over combinatorial structures. 
This algorithm successfully resolves the challenge posed by the combinatorial explosion of the search domain. 
It relies on two components:
The first is a novel acquisition algorithm that leverages techniques from convex optimization for scalability and efficiency.
Aside from effective handling of sparse data, the value of our model lies in its applicability to a wide range of combinatorial structures. 
We have demonstrated that our algorithm outperforms state-of-the-art methods from combinatorial optimization and machine learning.

Future work includes efficient optimization of other acquisition criteria, for example based on expected improvement or uncertainty reduction (e.g., see~\citet{chevalier2014fast,hernandez2014predictive}), and the development of tailored models for specific applications. 
For the latter, we anticipate a significant potential in the explicit modeling of combinatorial substructures that seem of relevance for a given task. 
For example, when optimizing over combinatorial structures such as graphical models, power grids and road networks, it seems promising to enrich the model by monomials that correspond to paths in the induced graph.
Another interesting direction is to employ a deep neural networks to learn useful representations for the regression. This technique would harmonize well with our acquisition function and complement the sparse parametric model proposed here for functions with moderate evaluation costs, as it would require more training data but also scale better to large numbers of samples.

\clearpage

\subsubsection*{Acknowledgement}
This work has been supported in part by the Air Force Office of Scientific Research (AFOSR) MURI on ``Managing multiple information sources of multi-physics systems," Program Officer Dr.\ Fariba Fahroo, Award Number FA9550-15-1-0038.

{
\bibliography{references}

\begin{thebibliography}{48}
\providecommand{\natexlab}[1]{#1}
\providecommand{\url}[1]{\texttt{#1}}
\expandafter\ifx\csname urlstyle\endcsname\relax
  \providecommand{\doi}[1]{doi: #1}\else
  \providecommand{\doi}{doi: \begingroup \urlstyle{rm}\Url}\fi

\bibitem[Arora \& Kale(2016)Arora and Kale]{arora2016combinatorial}
Arora, S. and Kale, S.
\newblock A combinatorial, primal-dual approach to semidefinite programs.
\newblock \emph{Journal of the ACM (JACM)}, 63\penalty0 (2):\penalty0 12, 2016.

\bibitem[Arora et~al.(2005)Arora, Berger, Elad, Kindler, and
  Safra]{arora2005non}
Arora, S., Berger, E., Elad, H., Kindler, G., and Safra, M.
\newblock On non-approximability for quadratic programs.
\newblock In \emph{46th Annual IEEE Symposium on Foundations of Computer
  Science (FOCS)}, pp.\  206--215, 2005.

\bibitem[Baptista et~al.(2018)Baptista, Marzouk, Willcox, and
  Peherstorfer]{baptista_2017}
Baptista, R., Marzouk, Y., Willcox, K., and Peherstorfer, B.
\newblock Optimal approximations of coupling in multidisciplinary models.
\newblock \emph{AIAA Journal}, 56\penalty0 (6):\penalty0 2412--2428, 2018.

\bibitem[Bergstra \& Bengio(2012)Bergstra and Bengio]{bb12}
Bergstra, J. and Bengio, Y.
\newblock Random search for hyper-parameter optimization.
\newblock \emph{Journal of Machine Learning Research}, 13:\penalty0 281--305,
  2012.

\bibitem[Bhattacharya et~al.(2016)Bhattacharya, Chakraborty, and
  Mallick]{bhattacharya_2016}
Bhattacharya, A., Chakraborty, A., and Mallick, B.~K.
\newblock Fast sampling with gaussian scale mixture priors in high-dimensional
  regression.
\newblock \emph{Biometrika}, pp.\  asw042, 2016.

\bibitem[Binois et~al.(2017)Binois, Ginsbourger, and
  Roustant]{binois2017choice}
Binois, M., Ginsbourger, D., and Roustant, O.
\newblock On the choice of the low-dimensional domain for global optimization
  via random embeddings.
\newblock \emph{arXiv preprint arXiv:1704.05318}, 2017.

\bibitem[Boyd \& Vandenberghe(2004)Boyd and Vandenberghe]{boyd2004convex}
Boyd, S. and Vandenberghe, L.
\newblock \emph{Convex optimization}.
\newblock Cambridge university press, 2004.

\bibitem[Brochu et~al.(2010)Brochu, Cora, and De~Freitas]{brochu2010tutorial}
Brochu, E., Cora, V.~M., and De~Freitas, N.
\newblock A tutorial on {Bayesian} optimization of expensive cost functions,
  with application to active user modeling and hierarchical reinforcement
  learning.
\newblock \emph{arXiv preprint arXiv:1012.2599}, 2010.

\bibitem[Carvalho et~al.(2010)Carvalho, Polson, and Scott]{carvalho_2010}
Carvalho, C.~M., Polson, N.~G., and Scott, J.~G.
\newblock The horseshoe estimator for sparse signals.
\newblock \emph{Biometrika}, 97\penalty0 (2):\penalty0 465--480, 2010.

\bibitem[Charikar \& Wirth(2004)Charikar and Wirth]{charikar_2004}
Charikar, M. and Wirth, A.
\newblock Maximizing quadratic programs: Extending grothendieck's inequality.
\newblock In \emph{Proc. of the 45th Annual IEEE Symposium on Foundations of
  Computer Science}, pp.\  54--60, 2004.

\bibitem[Chevalier et~al.(2014)Chevalier, Bect, Ginsbourger, Vazquez, Picheny,
  and Richet]{chevalier2014fast}
Chevalier, C., Bect, J., Ginsbourger, D., Vazquez, E., Picheny, V., and Richet,
  Y.
\newblock Fast parallel kriging-based stepwise uncertainty reduction with
  application to the identification of an excursion set.
\newblock \emph{Technometrics}, 56\penalty0 (4):\penalty0 455--465, 2014.

\bibitem[Del~Moral et~al.(2006)Del~Moral, Doucet, and Jasra]{del2006sequential}
Del~Moral, P., Doucet, A., and Jasra, A.
\newblock Sequential monte carlo samplers.
\newblock \emph{Journal of the Royal Statistical Society: Series B (Statistical
  Methodology)}, 68\penalty0 (3):\penalty0 411--436, 2006.

\bibitem[Dewancker et~al.(2016)Dewancker, McCourt, Clark, Hayes, Johnson, and
  Ke]{dewancker_2016}
Dewancker, I., McCourt, M., Clark, S., Hayes, P., Johnson, A., and Ke, G.
\newblock A stratified analysis of {Bayesian} optimization methods.
\newblock \emph{arXiv preprint arXiv:1603.09441}, 2016.

\bibitem[Garey \& Johnson(1979)Garey and Johnson]{garey2002computers}
Garey, M.~R. and Johnson, D.~S.
\newblock \emph{Computers and intractability}.
\newblock W. H. Freeman and Company, 1979.

\bibitem[Gelman et~al.(2013)Gelman, Carlin, Stern, and Dunson]{gelman_2014}
Gelman, A., Carlin, J.~B., Stern, H.~S., and Dunson, D.~B.
\newblock \emph{Bayesian data analysis}.
\newblock Chapman and Hall/CRC, 2013.

\bibitem[Golovin et~al.(2017)Golovin, Solnik, Moitra, Kochanski, Karro, and
  Sculley]{golovin2017google}
Golovin, D., Solnik, B., Moitra, S., Kochanski, G., Karro, J., and Sculley, D.
\newblock Google vizier: A service for black-box optimization.
\newblock In \emph{Proceedings of the 23rd ACM SIGKDD International Conference
  on Knowledge Discovery and Data Mining}, pp.\  1487--1495. ACM, 2017.

\bibitem[Hern{\'a}ndez-Lobato et~al.(2014)Hern{\'a}ndez-Lobato, Hoffman, and
  Ghahramani]{hernandez2014predictive}
Hern{\'a}ndez-Lobato, J.~M., Hoffman, M.~W., and Ghahramani, Z.
\newblock Predictive entropy search for efficient global optimization of
  black-box functions.
\newblock In \emph{Advances in {N}eural {I}nformation {P}rocessing {S}ystems},
  pp.\  918--926, 2014.

\bibitem[Hern{\'a}ndez-Lobato et~al.(2017)Hern{\'a}ndez-Lobato, Gonzalez, and
  Martinez-Cantin]{bayesopt_question}
Hern{\'a}ndez-Lobato, J.~M., Gonzalez, J., and Martinez-Cantin, R.
\newblock {NIPS} workshop on {B}ayesian optimization, 2017.
\newblock The problem is listed at \url{http://bayesopt.com/}. Last Accessed on
  02/07/18.

\bibitem[Hu et~al.(2010)Hu, Hu, Xu, Wang, and Cao]{hu_2010}
Hu, Y., Hu, J., Xu, Y., Wang, F., and Cao, R.~Z.
\newblock Contamination control in food supply chain.
\newblock In \emph{Proceedings of the 2010 Winter Simulation Conference}, pp.\
  2678--2681, 2010.
\newblock The source code is available at
  \url{http://simopt.org/wiki/index.php?title=Contamination_Control_Problem}.
  Last Accessed on 02/05/18.

\bibitem[Hutter(2009)]{hutter2009automated}
Hutter, F.
\newblock \emph{Automated configuration of algorithms for solving hard
  computational problems}.
\newblock PhD thesis, University of British Columbia, 2009.

\bibitem[Hutter \& Osborne(2013)Hutter and Osborne]{hutter2013kernel}
Hutter, F. and Osborne, M.~A.
\newblock A kernel for hierarchical parameter spaces.
\newblock \emph{arXiv preprint arXiv:1310.5738}, 2013.

\bibitem[Hutter et~al.(2010)Hutter, Hoos, and Leyton-Brown]{hutter_2010}
Hutter, F., Hoos, H., and Leyton-Brown, K.
\newblock Automated configuration of mixed integer programming solvers.
\newblock \emph{Integration of AI and OR Techniques in Constraint Programming
  for Combinatorial Optimization Problems}, pp.\  186--202, 2010.

\bibitem[Hutter et~al.(2011)Hutter, Hoos, and
  Leyton-Brown]{hutter2011sequential}
Hutter, F., Hoos, H.~H., and Leyton-Brown, K.
\newblock Sequential model-based optimization for general algorithm
  configuration.
\newblock In \emph{International Conference on Learning and Intelligent
  Optimization}, pp.\  507--523. Springer, 2011.

\bibitem[Jasa et~al.(2018)Jasa, Hwang, and Martins]{jasa_2017}
Jasa, J.~P., Hwang, J.~T., and Martins, J. R. R.~A.
\newblock Open-source coupled aerostructural optimization using python.
\newblock \emph{Structural and Multidisciplinary Optimization}, 57\penalty0
  (4):\penalty0 1815--1827, 2018.
\newblock The source code is available at
  \url{http://github.com/mdolab/OpenAeroStruct}. Last Accessed on 02/07/18.

\bibitem[Jenatton et~al.(2017)Jenatton, Archambeau, Gonz{\'a}lez, and
  Seeger]{jenatton2017bayesian}
Jenatton, R., Archambeau, C., Gonz{\'a}lez, J., and Seeger, M.
\newblock Bayesian optimization with tree-structured dependencies.
\newblock In \emph{International Conference on Machine Learning}, pp.\
  1655--1664, 2017.

\bibitem[Jones et~al.(1998)Jones, Schonlau, and Welch]{jsw98}
Jones, D.~R., Schonlau, M., and Welch, W.~J.
\newblock Efficient global optimization of expensive black-box functions.
\newblock \emph{Journal of Global Optimization}, 13\penalty0 (4):\penalty0
  455--492, 1998.

\bibitem[Kandasamy et~al.(2015)Kandasamy, Schneider, and
  P{\'o}czos]{kandasamy2015high}
Kandasamy, K., Schneider, J., and P{\'o}czos, B.
\newblock High dimensional {B}ayesian optimisation and bandits via additive
  models.
\newblock In \emph{International Conference on Machine Learning}, pp.\
  295--304, 2015.

\bibitem[Khanna et~al.(1998)Khanna, Motwani, Sudan, and Vazirani]{kmsv98}
Khanna, S., Motwani, R., Sudan, M., and Vazirani, U.~V.
\newblock On syntactic versus computational views of approximability.
\newblock \emph{{SIAM} Journal on Computing}, 28\penalty0 (1):\penalty0
  164--191, 1998.

\bibitem[Li et~al.(2016)Li, Kandasamy, P{\'o}czos, and Schneider]{li2016high}
Li, C.-L., Kandasamy, K., P{\'o}czos, B., and Schneider, J.
\newblock High dimensional bayesian optimization via restricted projection
  pursuit models.
\newblock In \emph{Artificial Intelligence and Statistics}, pp.\  884--892,
  2016.

\bibitem[Makalic \& Schmidt(2016)Makalic and Schmidt]{makalic_2016}
Makalic, E. and Schmidt, D.~F.
\newblock A simple sampler for the horseshoe estimator.
\newblock \emph{IEEE Signal Processing Letters}, 23\penalty0 (1):\penalty0
  179--182, 2016.

\bibitem[Mockus et~al.(1978)Mockus, Tiesis, and Zilinskas]{MoTiZi78}
Mockus, J., Tiesis, V., and Zilinskas, A.
\newblock {The application of Bayesian methods for seeking the extremum}.
\newblock In Dixon, L. C.~W. and Szego, G.~P. (eds.), \emph{Towards Global
  Optimisation}, volume~2, pp.\  117--129. Elsevier Science Ltd., North
  Holland, Amsterdam, 1978.

\bibitem[Negoescu et~al.(2011)Negoescu, Frazier, and Powell]{negoescu_2011}
Negoescu, D.~M., Frazier, P.~I., and Powell, W.~B.
\newblock The knowledge-gradient algorithm for sequencing experiments in drug
  discovery.
\newblock \emph{INFORMS Journal on Computing}, 23\penalty0 (3):\penalty0
  346--363, 2011.

\bibitem[Pankratov \& Borodin(2010)Pankratov and Borodin]{pankratov_2010}
Pankratov, D. and Borodin, A.
\newblock On the relative merits of simple local search methods for the
  {MAX-SAT} problem.
\newblock In \emph{{SAT} 2010}, pp.\  223--236, 2010.

\bibitem[Poloczek \& Williamson(2017)Poloczek and Williamson]{poloczek_2017}
Poloczek, M. and Williamson, D.~P.
\newblock An experimental evaluation of fast approximation algorithms for the
  maximum satisfiability problem.
\newblock \emph{ACM Journal of Experimental Algorithmics (JEA)}, 22:\penalty0
  1--18, 2017.

\bibitem[Russo et~al.(2017)Russo, Van~Roy, Kazerouni, and
  Osband]{russo2017tutorial}
Russo, D., Van~Roy, B., Kazerouni, A., and Osband, I.
\newblock A tutorial on {T}hompson sampling.
\newblock \emph{arXiv:1707.02038}, 2017.

\bibitem[Sch{\"a}fer(2013)]{schafer_2013}
Sch{\"a}fer, C.
\newblock Particle algorithms for optimization on binary spaces.
\newblock \emph{ACM Transactions on Modeling and Computer Simulation (TOMACS)},
  23\penalty0 (1):\penalty0 8, 2013.

\bibitem[Selman et~al.(1993)Selman, Kautz, and Cohen]{skc94}
Selman, B., Kautz, H.~A., and Cohen, B.
\newblock Local search strategies for satisfiability testing.
\newblock In \emph{Cliques, Coloring, and Satisfiability: Second DIMACS
  Implementation Challenge}, pp.\  521--532, 1993.

\bibitem[Shahriari et~al.(2016{\natexlab{a}})Shahriari, Bouchard-Cote, and
  Freitas]{shahriari2016unbounded}
Shahriari, B., Bouchard-Cote, A., and Freitas, N.
\newblock Unbounded bayesian optimization via regularization.
\newblock In \emph{Artificial Intelligence and Statistics}, pp.\  1168--1176,
  2016{\natexlab{a}}.

\bibitem[Shahriari et~al.(2016{\natexlab{b}})Shahriari, Swersky, Wang, Adams,
  and de~Freitas]{shahriari_2016}
Shahriari, B., Swersky, K., Wang, Z., Adams, R.~P., and de~Freitas, N.
\newblock Taking the human out of the loop: A review of {Bayesian}
  optimization.
\newblock \emph{Proceedings of the IEEE}, 104\penalty0 (1):\penalty0 148--175,
  2016{\natexlab{b}}.

\bibitem[Snoek et~al.(2012)Snoek, Larochelle, and Adams]{snoek_2012}
Snoek, J., Larochelle, H., and Adams, R.~P.
\newblock Practical {Bayesian} optimization of machine learning algorithms.
\newblock In \emph{Advances in {N}eural {I}nformation {P}rocessing {S}ystems},
  pp.\  2951--2959, 2012.

\bibitem[Spears(1993)]{spears_1993}
Spears, W.~M.
\newblock Simulated annealing for hard satisfiability problems.
\newblock In \emph{Cliques, Coloring and Satisfiability: Second DIMACS
  Implementation Challenge}, pp.\  533--558, 1993.

\bibitem[Steurer(2010)]{steurer2010fast}
Steurer, D.
\newblock Fast {SDP} algorithms for constraint satisfaction problems.
\newblock In \emph{Proceedings of the twenty-first annual ACM-SIAM symposium on
  Discrete Algorithms}, pp.\  684--697. SIAM, 2010.

\bibitem[Swersky et~al.(2014)Swersky, Duvenaud, Snoek, Hutter, and
  Osborne]{swersky2014raiders}
Swersky, K., Duvenaud, D., Snoek, J., Hutter, F., and Osborne, M.~A.
\newblock Raiders of the lost architecture: Kernels for bayesian optimization
  in conditional parameter spaces.
\newblock \emph{arXiv preprint arXiv:1409.4011}, 2014.

\bibitem[Thompson(1933)]{thompson1933likelihood}
Thompson, W.~R.
\newblock On the likelihood that one unknown probability exceeds another in
  view of the evidence of two samples.
\newblock \emph{Biometrika}, 25:\penalty0 285--294, 1933.

\bibitem[Thompson(1935)]{thompson1935theory}
Thompson, W.~R.
\newblock On the theory of apportionment.
\newblock \emph{American Journal of Mathematics}, 57\penalty0 (2):\penalty0
  450--456, 1935.

\bibitem[Wang et~al.(2016)Wang, Hutter, Zoghi, Matheson, and
  de~Feitas]{wang2016bayesian}
Wang, Z., Hutter, F., Zoghi, M., Matheson, D., and de~Feitas, N.
\newblock Bayesian optimization in a billion dimensions via random embeddings.
\newblock \emph{Journal of Artificial Intelligence Research}, 55:\penalty0
  361--387, 2016.

\bibitem[Wang et~al.(2017)Wang, Li, Jegelka, and Kohli]{wang2017batched}
Wang, Z., Li, C., Jegelka, S., and Kohli, P.
\newblock Batched high-dimensional {B}ayesian optimization via structural
  kernel learning.
\newblock \emph{arXiv preprint arXiv:1703.01973}, 2017.

\bibitem[Zhang et~al.(2015)Zhang, Sohn, Villegas, Pan, and Lee]{zhang_2015}
Zhang, Y., Sohn, K., Villegas, R., Pan, G., and Lee, H.
\newblock Improving object detection with deep convolutional networks via
  {Bayesian} optimization and structured prediction.
\newblock In \emph{The IEEE Conference on Computer Vision and Pattern
  Recognition (CVPR)}, June 2015.

\end{thebibliography}
\bibliographystyle{icml2018}
}

\appendix

\cleardoublepage
\newpage

\begin{center}
{ \LARGE
\bf 
Bayesian Optimization of Combinatorial Structures
}
\vspace{2mm}

{\textit{Supplementary Material}}

\end{center}

\section{General Form of \CSS}
\label{section_general_form_CSS}
In this section we describe the general form of \CSS that handles models of order larger than two as well as categorical and integer-valued variables.

So far, we have focused our presentation on binary variables, i.e., $\domain = \{0,1\}^d$ or equivalently, $\domain = \{-1,+1\}^d$, that allow efficient encodings of many combinatorial structures as demonstrated above.

We begin with a description of how to incorporate categorical variables into our statistical model.
Let~$\cal I$ denote the indices of categorical variables. Consider a categorical variable~$x_i$ with~$i \in {\cal I}$ that takes values in~$\domain_i = \{e^i_1,\ldots,e^i_{m_i}\}$. 
We introduce~$m_i$ new binary variables~$x_{ij}$ with~$x_{ij} = 1$ if~$x_i = e^i_j$ and~$x_{ij} = 0$ otherwise.
Note that~$\sum_j x_{ij} = 1$ for all~$i \in {\cal I}$ since the variable takes exactly one value, and the dimensionality of the binary variables increases from~$d$ to~$d - |{\cal I}| + \sum_{i \in {\cal I}} m_i$.

\CSS uses the sparse Bayesian linear regression model for binary variables proposed in Sect.~\ref{sec:statistical_model} and samples $\balpha_t$ in each iteration~$t$. 
When searching for the next~$x^{(t)}$ that optimizes the objective value for~$\balpha_t$, we need to exert additional care: observe that running \SA to optimize the binary variables might result in a solution that selects more than one element in~$\domain_i$, or none at all, and therefore would not correspond to a feasible assignment to the categorical variable~$x_i$.

Instead, we undo the above expansion: \SA operates on $d$-tuples~$x$ where each~$x_i$ with~$i \in {\cal I}$ takes values in its original domain~$\domain_i$. Then the neighborhood~$N(x)$ of any tuple~$x$ is given by all vectors where at most one variable differs in its assignment. 
To evaluate~$f_{\balpha_t}(x) + {\cal P}(x)$, we leverage this correspondence between categorical variables and their `binary representation'.

Note that integer-valued variables can be handled naturally by the regression model.
For the optimization of the acquisition criterion, simulated annealing uses the same definition of the neighborhood~$N(x)$ as in the case of categorical variables.

Next we show how to extend the \CSS algorithm to models that contain monomials of length larger than two.
In this case we have $$f_\alpha(x) = \sum_{S \in 2^\domain} \alpha_S \prod_{i \in S}x_i,$$ where~$2^\domain$ denotes the power set of the domain and~$\alpha_S$ is a real-valued coefficient.

Following the description of \CSS for second-order models, the regression model is obtained by applying the sparsity-inducing prior described in Sect.~\ref{sec:statistical_model}.
Then, in each iteration~$t$, we sample~$\balpha_t$ from the posterior over the regression coefficients and now search for an~$x^{(t)}$ that approximates
\begin{equation*}
\max_{x \in \domain} \sum_{S \in 2^\domain} \alpha_S \prod_{i \in S}x_i  - \lambda {\cal P}(x).
\end{equation*}
Since we can evaluate the objective value $f_{\balpha_t}(x) - \lambda {\cal P}(x)$ at any~$x$ efficiently, we may use simulated annealing again to find an approximate solution to the acquisition criterion.

\section{Evaluation of Higher Order Models}
\label{section_experiments_higher_order_models}
We point out that the problems studied in Sect.~\ref{sec:sparse_ising}, \ref{sec:cont_control}, and \ref{sec:aerospace_test} have natural interactions of order higher than two between the elements that we optimize over. 
To highlight these interactions, we measure the number of regression coefficients that have significant weight (i.e., values $|\alpha_{i}| \geq 0.1$) with the sparse regression model of different orders.

As an example, we fit the model using 100 samples from a random instance of the Ising model presented in Sect.~\ref{sec:sparse_ising}. 
Typically, four out of 24 linear terms, 28 out of 300 second-order terms, and 167 out of 2048 third-order terms have value of at least~$0.1$.  
Here we note the importance of the sparsity-inducing prior to promote a small number of parameters in order to reduce the variance in the model predictions (cp.\ Sect.~\ref{sec:statistical_model}).

We also examine how \CSS performs when equipped with a statistical model of higher order. Our implementation follows Sect.~\ref{sec:statistical_model} and uses simulated annealing to search for an optimizer of the acquisition criterion as described in Sect.~\ref{section_general_form_CSS} and in Sect.~\ref{section_CSS_SA}. 

\begin{figure}[!h]
\centering
\includegraphics[width=0.9\linewidth]{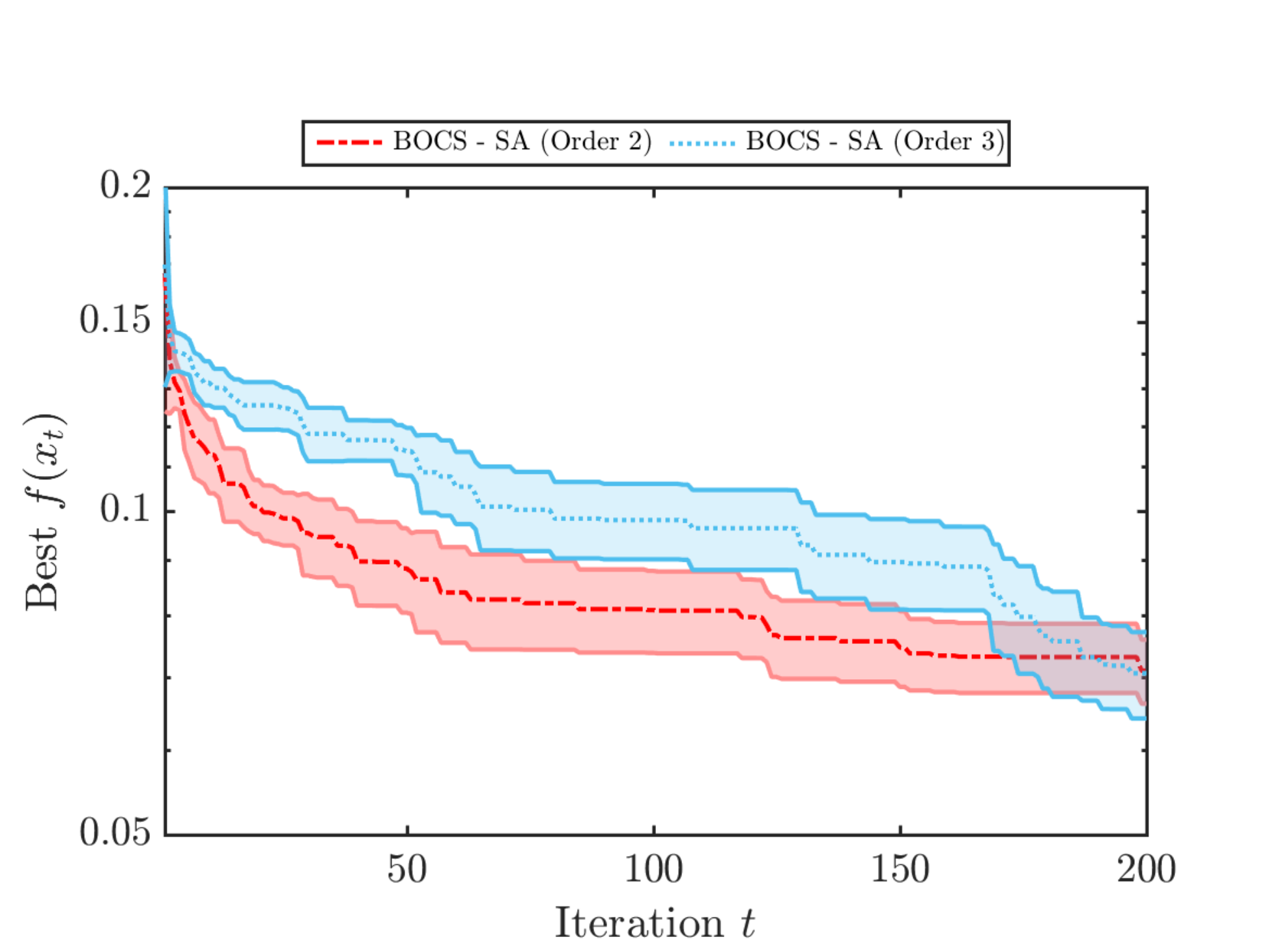}
\vspace{-10pt}
\caption{Performance of \CSSSA on the aero-structural benchmark for $\lambda = 10^{-2}$ with second and third-order statistical models. As the number of samples increases, \CSSSA with the third-order model achieves better results.
\label{fig:aerostruct_higherorder}}
\end{figure}

Fig.~\ref{fig:aerostruct_higherorder} compares the performances of the \CSSSA algorithm on the aero-structural benchmark with a second and third-order model. 
The second-order model has a lower number of coefficients that can be estimated with lower variance given few training samples.
On the other hand, a statistical model of higher order is able to capture more interactions between the active coupling variables but may require more samples for a sufficient model fit.
Thus, it is not surprising that \CSSSA performs better initially with the second-order model. As the number of samples grows larger, the third order model obtains better results.

\begin{figure}[!h]
\centering
\includegraphics[width=0.9\linewidth]{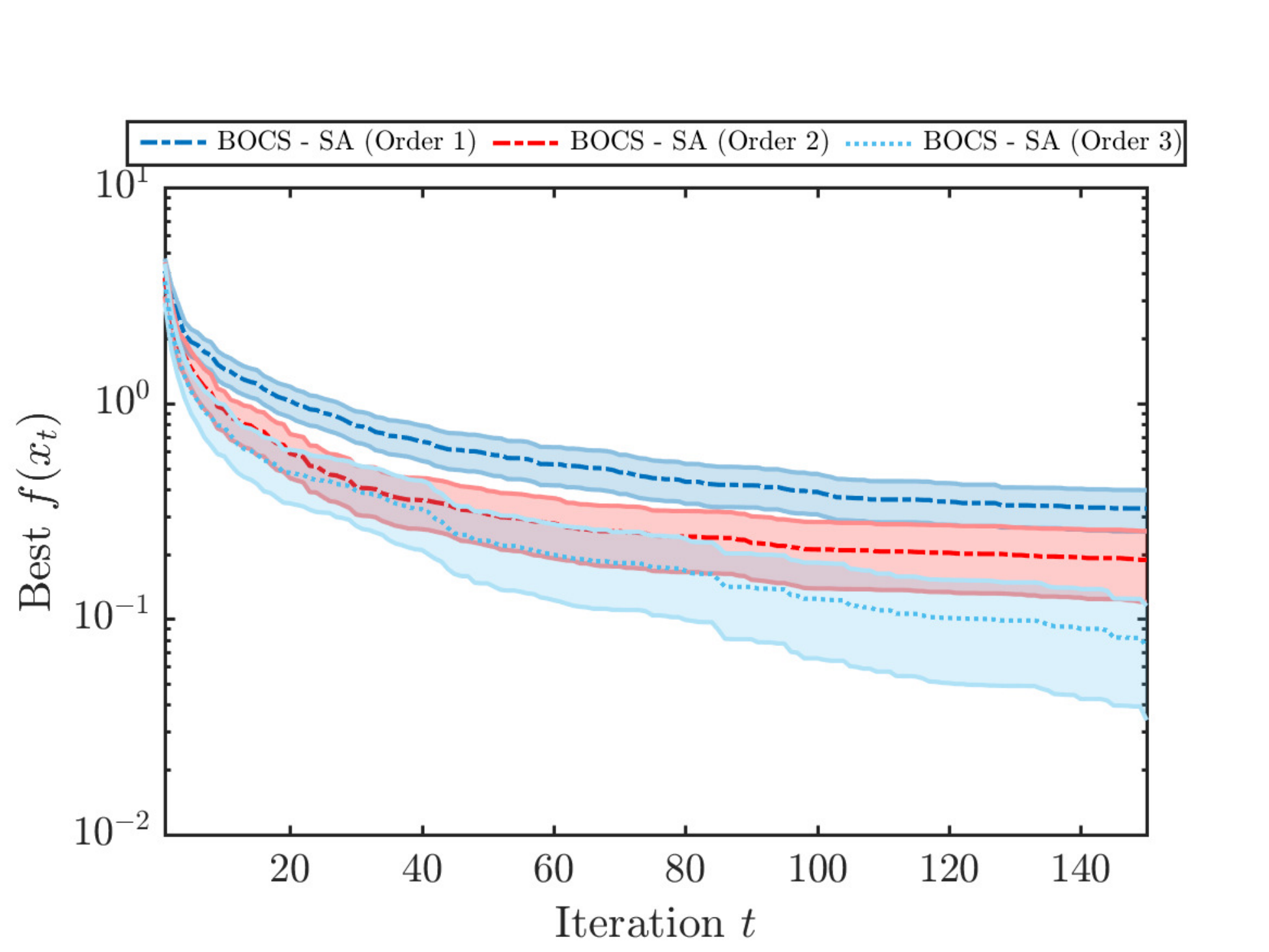}
\vspace{-10pt}
\caption{Performance of \CSSSA for the Ising model with $\lambda {=} 0$. After 150 iterations, \CSSSA with the third order model performs better on average. \label{fig:isinghighorder_l0}}
\end{figure}

\begin{figure}[!h]
\centering
\includegraphics[width=0.9\linewidth]{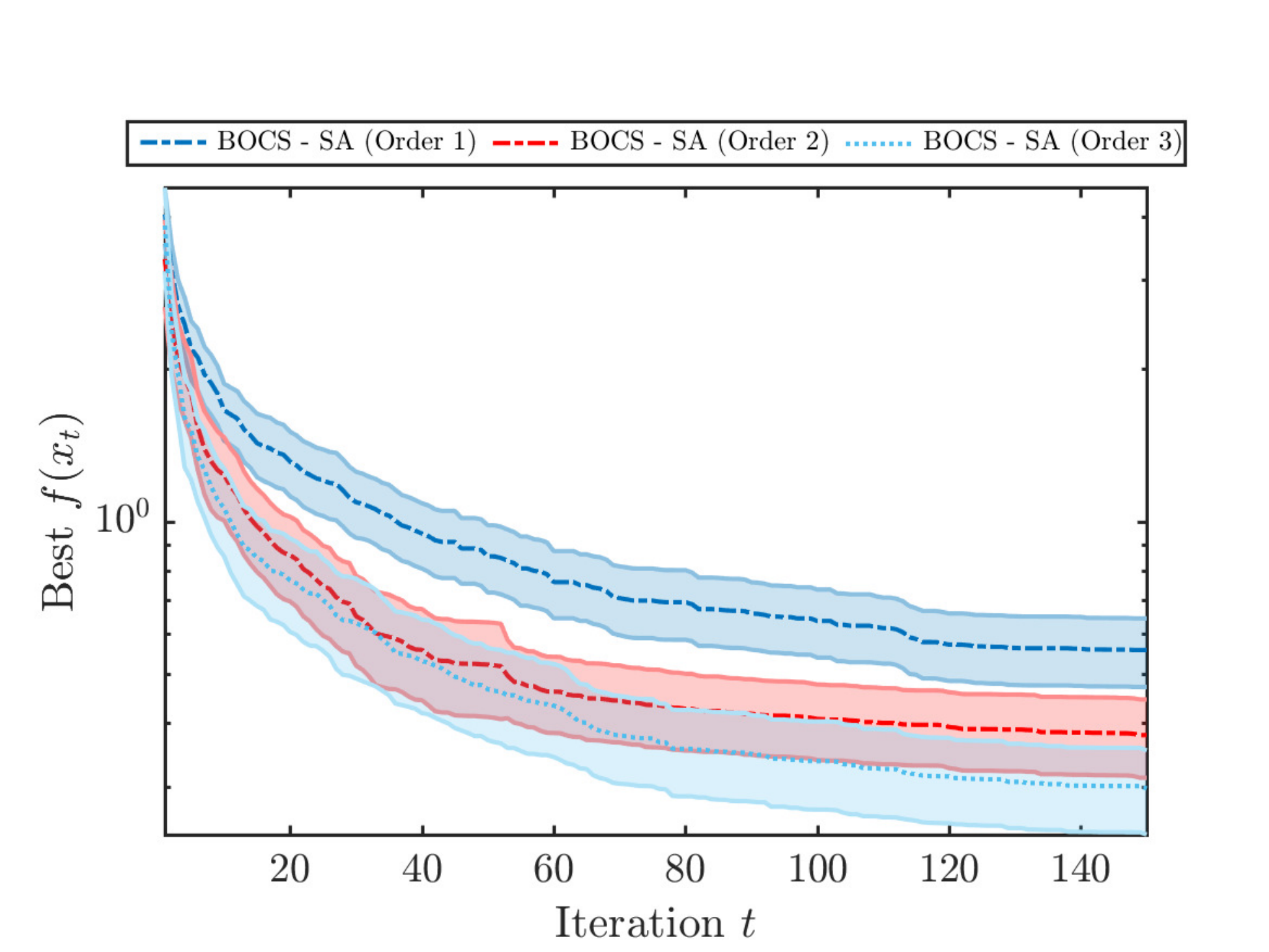}
\vspace{-10pt}
\caption{Performance of \CSSSA on the Ising model with $\lambda {=} 10^{-2}$. After 150 iterations, the third order models typically leads to better results. \label{fig:isinghighorder_lem2}}
\end{figure}

We also evaluate \CSSSA with higher order models for the Ising benchmark presented in Sect.~\ref{sec:sparse_ising}. Fig.~\ref{fig:isinghighorder_l0} contrasts the performances of the \CSSSA algorithm for $\lambda = 0$ with a first order, a second-order, and a third-order statistical model. All results are averaged over 100 instances of the Ising model.
Fig.~\ref{fig:isinghighorder_lem2} summarizes the results for~$\lambda = 10^{-2}$. 
Interestingly, the third-order model already performs similarly to the second-order model for this problem with a smaller number of data points, although it exhibits a larger variability in the performance. 

These results provide numerical evidence that a second-order model provides a good trade-off of model expressiveness and accuracy for these problems when data is limited.

\section{Wall-clock Time Performance}
\label{section_wallclock_times}
In this section, we compare the wall-clock times required by \CSS and \EI for the Ising benchmark presented in Sect.~\ref{sec:sparse_ising}. 
The wall-clock time is computed as the first time each instance of the algorithm reaches an objective value of $0.01$ for $\lambda = 0$, and $10^{-4}$. The average results over~$100$ runs of each algorithm and the $95\%$ confidence intervals are presented in Table~\ref{tab:wall_clock_time_ising}.

\begin{table}[!ht]
	\caption{Wall-clock time required by the algorithms presented in Sect.~\ref{sec:algorithms} for the 24-dimensional Ising model benchmark. \CSSSA and \CSSSDP are considerably more efficient than~\EI. 
		\label{tab:wall_clock_time_ising}}
	\vskip 0.05in
	\begin{center}
		\begin{small}
			\begin{sc}
				\begin{tabular}{lccc}
					
					\toprule
					$\lambda$   & \EI & \CSSSA & \CSSSDP \\
					
					\midrule
					$0$   	    & $404.2 \pm 49.1$ & $\textbf{45.8} \pm 12.4$ & $115.6 \pm 18.1$ \\
					$10^{-4}$   & $412.9 \pm 55.8$ & $\textbf{62.1} \pm 16.5$ & $104.6 \pm 16.7$ \\
					\bottomrule
				\end{tabular}
			\end{sc}
		\end{small}
	\end{center}
	\vskip -0.1in
\end{table}

The results demonstrate that~\CSS is considerably faster than~\EI. \CSSSA is at least seven times faster, while \CSSSDP is still four times faster.  Even for a problem with 24 binary variables, the cost of finding an optimizer of the acquisition function is prohibitively large for~\EI.

\section{Descriptions of Benchmark Problems}

In this section, we provide more details on the benchmark problems studied in Sect.~\ref{sec:num_results}.

\subsection{Sparsification of Ising Models}

To evaluate the objective function for the Ising model (see Sect.~\ref{sec:sparse_ising}), we compute the KL divergence between models $p(z)$ and $q_{x}(z)$, that are defined by their interaction matrices $J^{p}$ and $J^{q}$, respectively. To do so, we pre-compute the second moments of the random variables in the original model given by $\mathbb{E}_{p}[z_{i}z_{j}]$ and use these together with the differences in the interaction matrices to evaluate the first term in the KL divergence. The second term in the objective is given by the log difference of the partition functions, $\log(Z_{q}/Z_{p})$, where $Z_{p}$ is constant for each $x$ and only needs to be evaluated once. $Z_{q}$ is the normalizing constant for the approximating distribution and is given by
\begin{equation}
Z_{q} = \sum_{z \in \{-1,1\}^{n}} \exp(z^{T}J^{q}z).
\end{equation}
In this work we do not restrict the class of distributions to be defined over subgraphs $\mathcal{G}^{q}$ whose normalizing constants can be computed efficiently (e.g., mean-field approximations). Therefore, in general, computing $Z_{q}$ requires summing an exponential number of terms with respect to $n$, making this term and the KL divergence expensive to evaluate.

Furthermore, the objective function above only measures the distance between $p(z)$ and an approximating distribution defined over a subgraph while still using the same parameters, i.e., if~$J^{q}_{ij}$ is non-zero then it has the same value as the corresponding entry in~$J^{p}$.

\subsection{Contamination Control}

The objective in this problem is to minimize the cost of prevention efforts while asserting that the contamination level does not exceed certain thresholds with sufficiently high probability.
These latter constraints are evaluated by running $T$ Monte Carlo simulations and counting the number of runs that exceed the specified upper limits for the contamination. 
Each set of Monte Carlo runs defines an instance of the objective function in Eq.~(\ref{eq:obj_foodcontrol}) that we optimize with respect to~$x$ using the various optimization algorithms presented in Sect.~\ref{sec:algorithms}. 
In our studies we followed the recommended problem parameters that are provided by the \texttt{SimOpt} Library \cite{hu_2010}.

\subsection{Aero-structural Multi-Component Problem}

The \emph{OpenAeroStruct} model developed by~\cite{jasa_2017} computes three output variables for each set of random input parameters. In this study, our objective is to evaluate the change in the probability distribution of these outputs for each set of active coupling variables between the components of the computational model, $x$. 
We denote the distribution of the outputs in the reference model by $\pi_{\mathbf{y}}$ (i.e., with all active coupling variables) and the decoupled model by $\pi_{\mathbf{y}}^{x}$.
The difference in these probability distributions is measured by the KL divergence and is denoted by $D_{KL}(\pi_{\mathbf{y}}||\pi_{\mathbf{y}}^{x})$. 

While the KL divergence can be estimated to arbitrary accuracy with Monte Carlo simulation and density estimation techniques, in this study we follow \citet{baptista_2017} and rely on an approximation of the objective. 
This approximation linearizes the components of the model and propagates the uncertainty in the Gaussian distributed input variables to characterize the Gaussian distribution for the outputs. By repeating this process for the reference and decoupled models, an estimate for the KL divergence can be computed in closed form between the two multivariate Gaussian distributions. However, the linearization process still requires computing gradients with respect to high-dimensional internal state variables within the model and is thus computationally expensive.

For more information on how to evaluate the approximate KL divergence as well as its numerical performance in practice for several engineering problems, we refer the reader to~\citet{baptista_2017}.

\section{Maximum Likelihood Estimate for the Regression Coefficients}
\label{section_mle_model}
In Sect.~\ref{sec:statistical_model} we proposed a Bayesian treatment of the regression coefficients~$\balpha$ in Eq.~\eqref{eq:stat_model}. 
Here we discuss an alternative approach based on a point-estimate, e.g., a maximum likelihood estimate (MLE).
Suppose that we have observed $(x^{(i)},f(x^{(i)}))$ for $i=1,\dots,N$.
The maximum likelihood estimator assumes that the discrepancy between $f(x)$ and the statistical model is represented with an additive error. This error is supposed to follow a normal distribution with mean zero and known finite variance $\sigma^2$. 
To compute this estimator, we stack the~$p$ predictors of all~$N$ samples to obtain $\mathbf{X} \in \{0,1\}^{N \times p}$. 
Then the regression coefficients~$\balpha$ are obtained by the least-squares estimator
\begin{equation} 
\label{eq:mle_estimate}
\balpha_{MLE} = (\mathbf{X}^{T}\mathbf{X})^{-1}(\mathbf{X}^{T}\mathbf{y}),
\end{equation}
where~$\mathbf{y} \in \mathbb{R}^{N}$ is the vector of function evaluations. 

While the parameters of this model can be efficiently estimated for a small number of evaluations, the MLE only provides a uniform estimate of $\sigma^2$ for the variance of the coefficients. On the other hand, the Bayesian models described in Section~\ref{sec:statistical_model} better characterize the joint uncertainty of all parameters $\balpha$ and $\sigma^2$ in order to capture the discrepancy of the generative model. \CSS leverages this uncertainty by sampling from the posterior distribution over the coefficients. This sampling allows \CSS to better explore the combinatorial space of models and find a global optimum of the objective function. This is also contrasted with using a variance of $\sigma^2$ to sample the coefficients independently, which may lead to uninformative models that do not account for the correlation between coefficients that is captured by the Bayesian models. Furthermore, we note that only using the MLE coefficients from~\eqref{eq:mle_estimate} in \CSS often results in purely exploitative behavior that fails to find the global optimum, as observed in Fig.~\ref{fig:quad_utility_gap_alpha100_lambda1}.

\section{Validation of the Regression Models}
\label{section_validation_models}
We now validate models of order two proposed in Sect.~\ref{sec:statistical_model} and Sect.~\ref{section_mle_model} for each benchmark considered in Sect.~\ref{sec:num_results}. The figures compare the statistical models based on the maximum likelihood estimate, standard Bayesian linear regression and sparse Bayesian linear regression based on the sparsity-inducing prior introduced in Sect.~\ref{sec:sparse_blr}. The standard Bayesian linear regression model supposes a joint prior for the parameters of $P(\boldsymbol{\alpha},\sigma^2) = P(\boldsymbol{\alpha}|\sigma^2)P(\sigma^2)$, where $\balpha|\sigma^2 \sim \mathcal{N}(\mu_{\alpha},\sigma^2\Sigma_{\alpha})$ and $\sigma^2 \sim IG(a,b)$ have a normal and inverse-gamma distribution respectively. Given the same data model as in Sect.~\ref{sec:sparse_blr}, the joint posterior of~$\balpha$ and~$\sigma^2$ has a normal-inverse-gamma form.

We compare the average absolute approximation error of these three regression models on a test set of $M = 50$ points, varying the number of training points. 

\subsection{Validation on Binary Quadratic Programming.} 
\begin{figure}[!ht]
\centering
\includegraphics[width=0.8\linewidth]{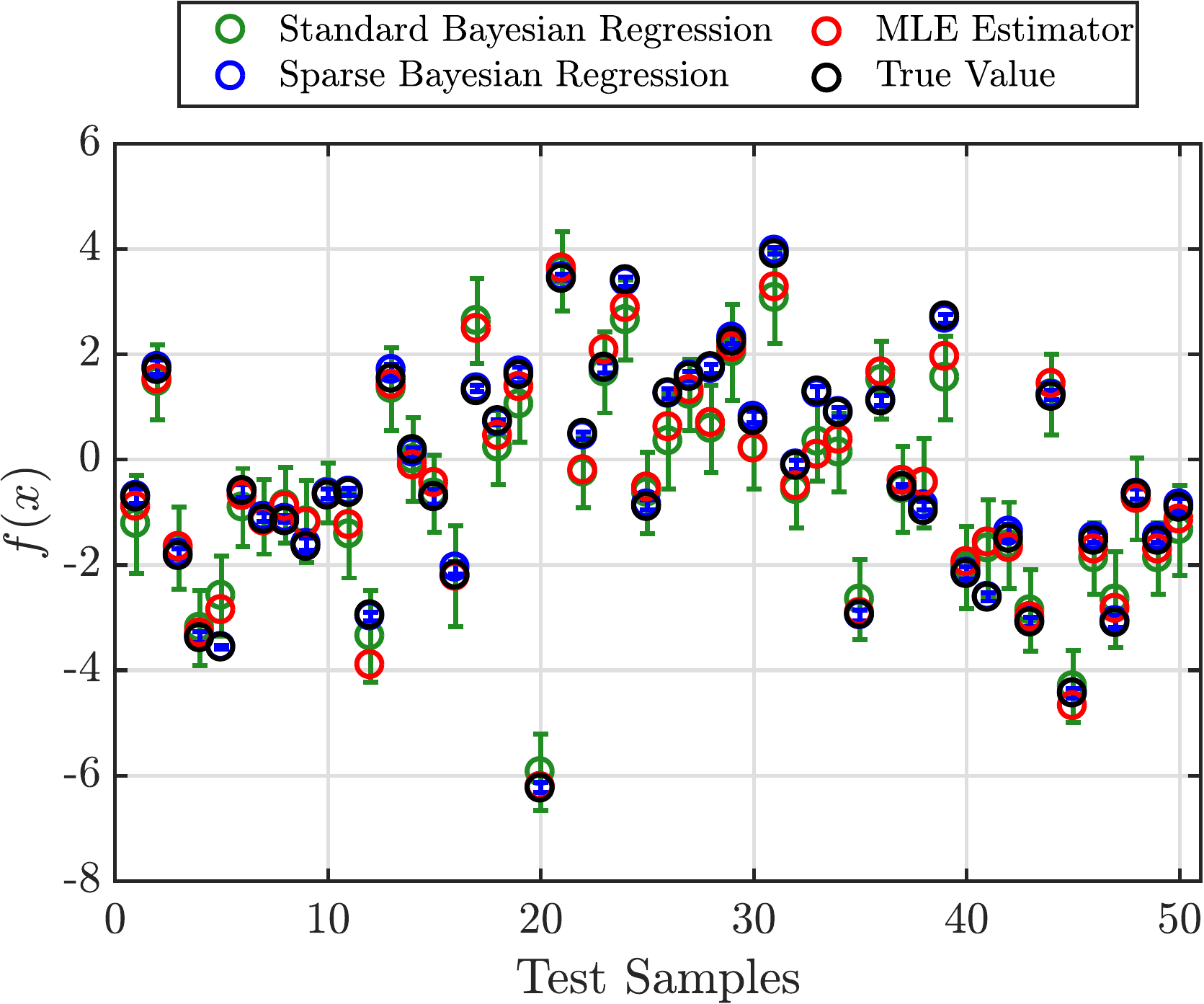}
\vspace{-10pt}
\caption{Test error of quadratic test problem ($L_{c} = 1$) for different $\balpha$ estimators. The Bayesian regression with a sparsity-inducing prior (blue) performs better than the Bayesian linear regression model (green) and the MLE estimator (red). \label{fig:quad_crossval}}
\end{figure}

We first evaluate the regression models on an instance of the test function from Sect.~\ref{sec:quadratic_test}, using a set of~$N = 40$ training samples.
Fig.~\ref{fig:quad_crossval} depicts: the true function values (black), the predictions of the MLE estimator (red), and the mean and standard deviation of the Bayesian linear regression model (green) and of the sparse regression model (blue). 
The regression model with the sparsity-inducing prior (blue) achieves the best prediction of the true values (black).
This figure empirically demonstrates the ability of the model to accurately capture the effect of binary coupling between input variables.
Although the MLE also provides good estimates, as we discuss in Sect.~\ref{sec:num_results}, the performance of the Bayesian optimization process is drastically impaired when using the MLE estimator instead of samples from the posterior of the regression coefficients, since the uncertainty in the model is not reflected in the former.

\begin{figure}[!ht]
\centering
\includegraphics[width=0.8\linewidth]{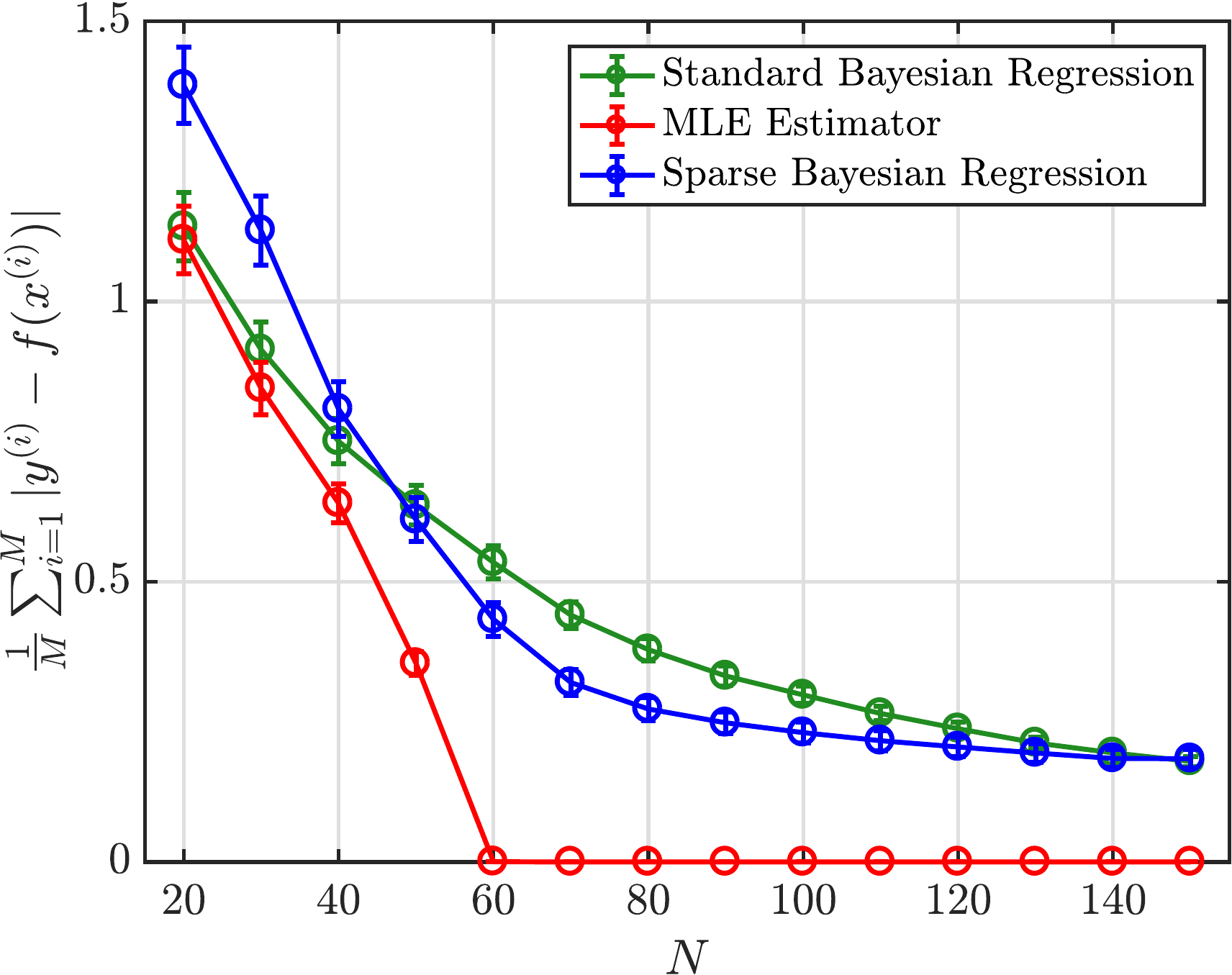}
\vspace{-10pt}
\caption{Test error of quadratic test problem with increasing size of training set. Standard Bayesian regression (green) and sparse regression (blue) perform similarly as $N$ increases for the $d=10$ quadratic test problem. \label{fig:quad_test_error}}
\end{figure}

We now compare the average test error of $M = 50$ points with an increasing size of the training set in Fig.~\ref{fig:quad_test_error}. The results are averaged over $30$ random instances of the binary quadratic problem (BQP) with $L_{c} \in [1,10,100]$. As $N$ increases, the test error is converging for all estimators. We note that for a quadratic objective function, the quadratic model $f_{\balpha}(x)$ closely interpolates the function with a sufficient number of training points, resulting in low test error for the MLE estimator. We note that for this lower $d=10$-dimensional test problem, standard Bayesian linear regression (green) resulted in similar accuracy as the sparse estimator (blue). 

\subsection{Validation on the Ising Problem.}

\begin{figure}[!ht]
\centering
\includegraphics[width=0.8\linewidth]{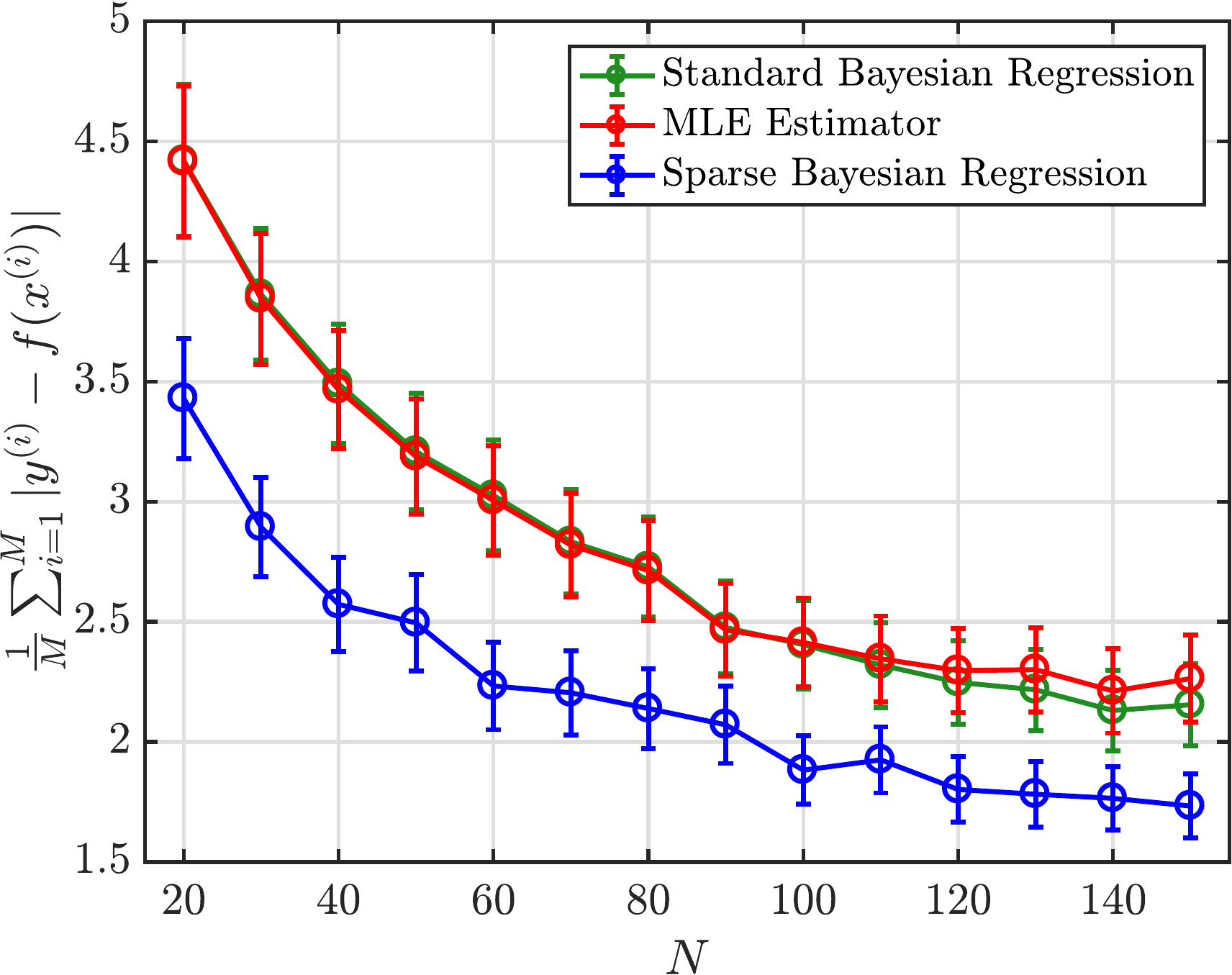}
\vspace{-10pt}
\caption{Test error of Ising model for different $\balpha$ estimators. The sparse estimator (blue) provides lower errors on the test set than standard Bayesian linear regression (red). \label{fig:ising_test_error}}
\end{figure}

For the Ising model with $d = 24$ edges, we examine the test error of $M = 50$ points with an increasing size of the training set; see Fig.~\ref{fig:ising_test_error}. 
The results are averaged over $10$ models with randomly drawn edge weights as discussed in Section~\ref{sec:sparse_ising} and the $95\%$ confidence intervals of the mean error are also reported in the error bars. As compared to the results for the BQP, the sparse estimator provides lower test errors for this higher-dimensional problem, warranting its use over Bayesian linear regression in the \CSS algorithm. This reduction in test error can be attributed to the shrinkage of coefficients with near-zero values from the sparsity-inducing prior~\cite{carvalho_2010}. 

\subsection{Validation on the Contamination Control Problem.}

\begin{figure}[!ht]
\centering
\includegraphics[width=0.8\linewidth]{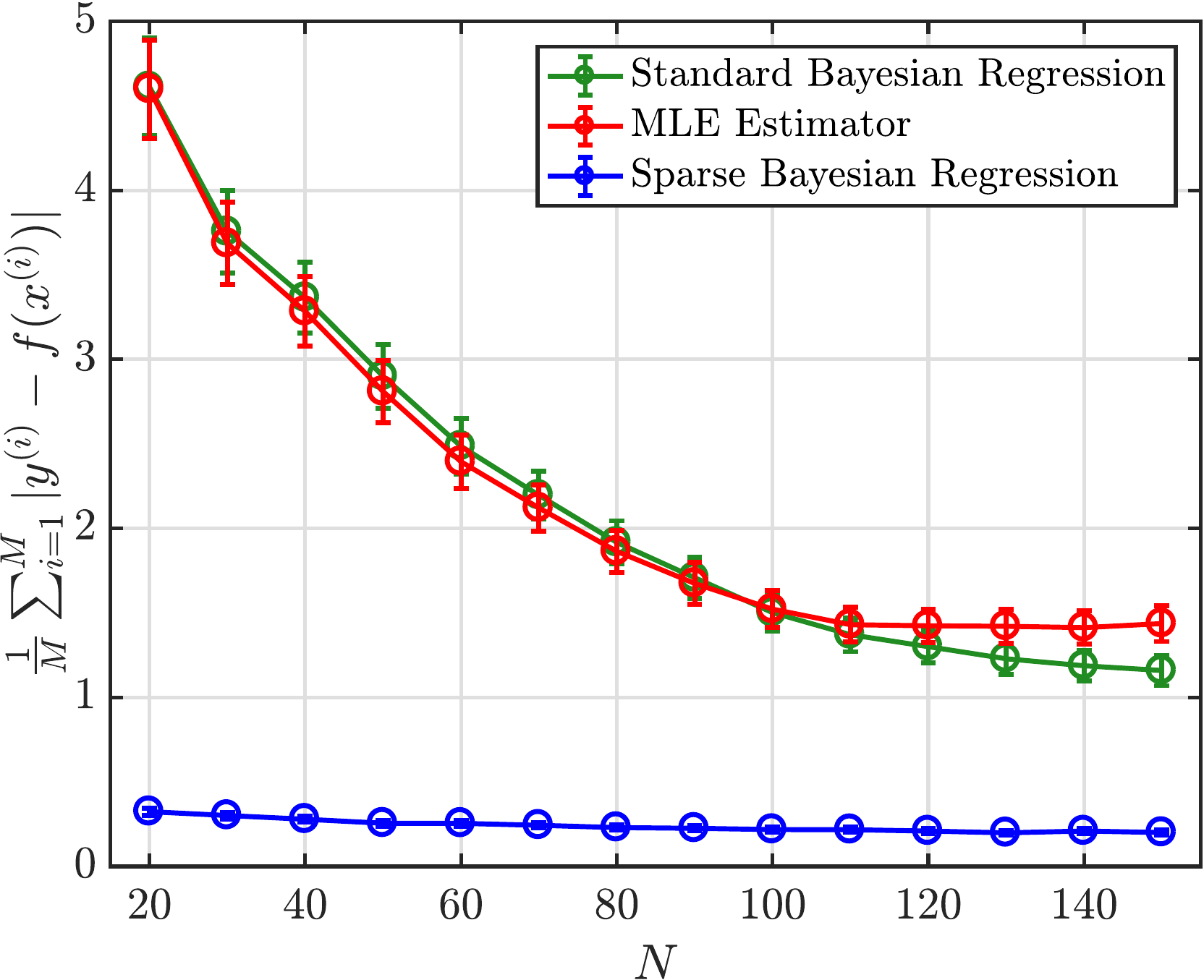}
\vspace{-10pt}
\caption{Test error of contamination control problem for different $\balpha$ estimators. The model based on the sparsity-inducing prior (blue) provides the best performance for approximating the objective. \label{fig:cont_test_error}}
\end{figure}

The test error with increasing training set size is plotted in Fig.~\ref{fig:cont_test_error} for the contamination control problem with $d = 20$ stages and $T = 10^{3}$ Monte Carlo samples for approximating the probability in the objective. 

With increasing~$N$, the variance in the values of all estimated coefficients decreases, which results in lower test set error as observed for the MLE and Bayesian linear regression. A similar behavior is also seen for the sparse estimator with a large reduction in the error offset for small values of $N$. 
This suggests that the objective can be well approximated by the model described in Section~\ref{sec:statistical_model} with a sparse set of interaction terms. 
As a result, the sparsity-inducing prior learns the set of non-zero terms and the test set error is dominated by the variance of the few remaining terms. 
It seems advantageous for \CSS to have a more accurate model based on this sparse prior when~$N$ is small.

\subsection{Validation on the Aero-structural Problem.}

\begin{figure}[!ht]
\centering
\includegraphics[width=0.8\linewidth]{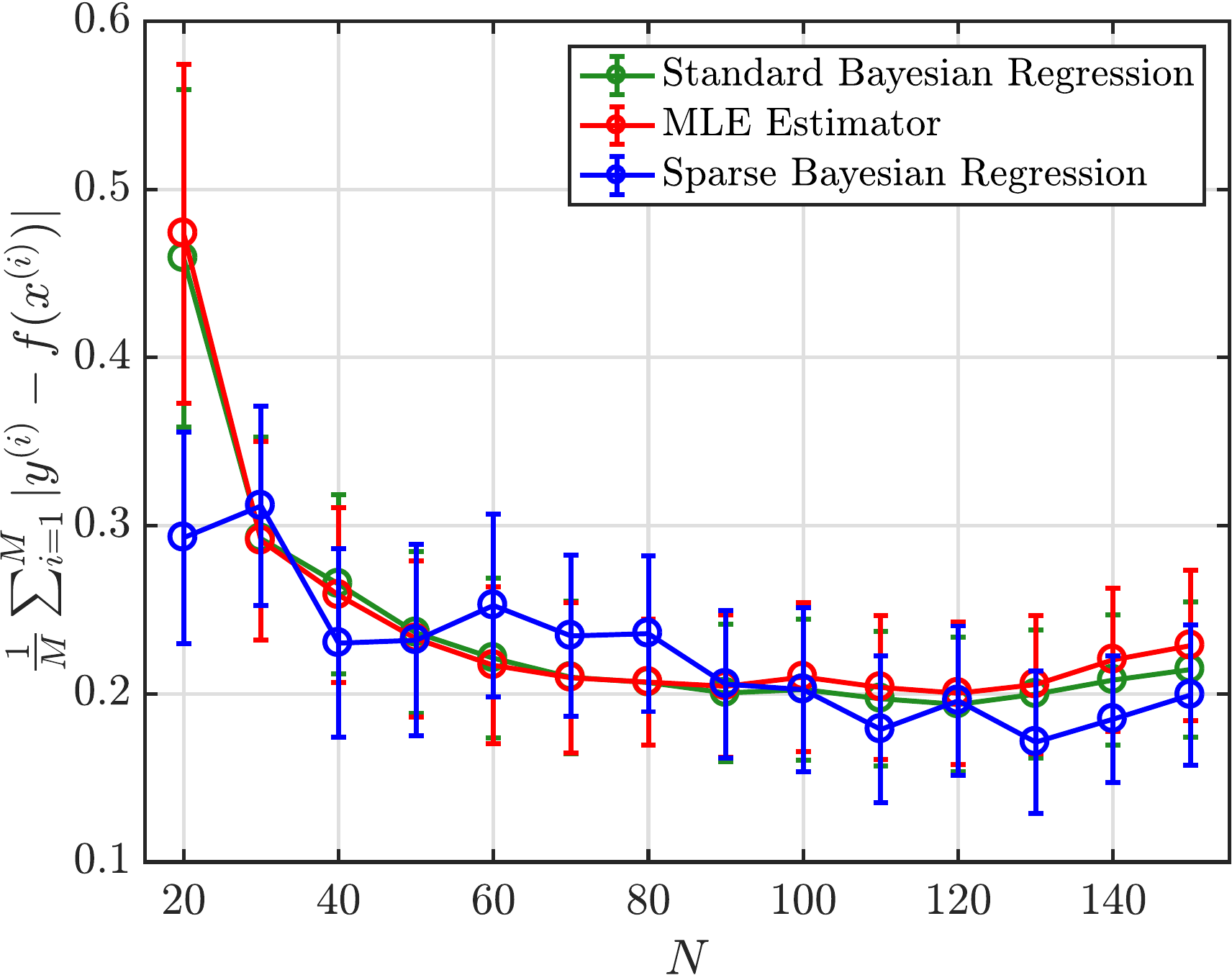}
\vspace{-10pt}
\caption{Test error of aero-structural problem for different $\balpha$ estimators. The MLE (red), standard Bayesian linear regression (green) and sparse linear regression (blue) produce similar test error results. \label{fig:aero_test_error}}
\end{figure}

For the aero-structural problem in Section~\ref{sec:aerospace_test}, the average absolute test set error for $M = 50$ samples is presented in Fig.~\ref{fig:aero_test_error}. With an increasing number of training samples, this problem has similar performance for the four different estimators. We note that for large $N$, the test set error of the $\balpha$ estimators for this problem begin to plateau with more training samples. This is an indication of the bias present in the statistical model of order two, and that it may be advantageous to use a higher order model to approximate the objective within \CSS, as observed in Fig.~\ref{fig:aerostruct_higherorder} with greater $N$. While the order two model may be computationally efficient, future work will address adaptive switching to a higher order when there are enough training samples to estimate its parameters with low variance.

\end{document}